%% file: main.tex
\documentclass[conference]{IEEEtran}
\usepackage{times}
\usepackage{amsmath}
\usepackage{amssymb}
\usepackage{graphicx}
\usepackage{mwe}
\usepackage{xcolor}

\usepackage{amsthm}
\newtheorem{definition}{Definition}

\newtheorem{lemma}{Lemma}
\newtheorem{remark}{Remark}
\newtheorem{assumption}{Assumption}
\newtheorem{example}{Example}

\usepackage[numbers]{natbib}
\usepackage{multicol}
\usepackage[bookmarks=true]{hyperref}
\newcommand{\hquad}{\hspace{0.5em}}

\begin{document}

\title{Collaborative Object Transportation in Space \\ via Impact Interactions}



\author{\authorblockN{Joris Verhagen}
\authorblockA{School of Electrical Engineering and Computer Science\\
KTH Royal Institute of Technology\\
Stockholm, Sweden\\
Email: jorisv@kth.se}
\and
\authorblockN{Jana Tumova}
\authorblockA{School of Electrical Engineering and Computer Science\\
KTH Royal Institute of Technology\\
Stockholm, Sweden\\
Email: tumova@kth.se}}


%

\maketitle

\begin{abstract}

We present a planning and control approach for collaborative transportation of objects in space by a team of robots. Object and robots in microgravity environments are not subject to friction but are instead free floating. This property is key to how we approach the transportation problem: the passive objects are controlled by impact interactions with the controlled robots.
In particular, given a high-level Signal Temporal Logic (STL) specification of the transportation task, we synthesize motion plans for the robots to maximize the specification satisfaction in terms of spatial STL robustness.
Given that the physical impact interactions are complex and hard to model precisely, we also present an alternative formulation maximizing the permissible uncertainty in a simplified kinematic impact model.
We define the full planning and control stack required to solve the object transportation problem; an offline planner, an online replanner, and a low-level model-predictive control scheme for each of the robots. 
We show the method in a high-fidelity simulator for a variety of scenarios and present experimental validation of 2-robot, 1-object scenarios on a freeflyer platform.

Code and Videos at \url{https://joris997.github.io/impact_stl/}

\end{abstract}

\IEEEpeerreviewmaketitle

\input{1_Introduction}

\input{2_Preliminaries}

\input{3_Problem}

\input{4_Method}

\input{5_Implementation}

\input{6_Results}

\input{7_Limitations}

\input{8_Conclusion}


\bibliographystyle{plainnat}
\bibliography{references}

\clearpage
\appendix
\input{Appendix}

\end{document}

%% file: 1_Introduction.tex
\section{Introduction}
\label{sec:introduction}


The challenge of object transportation with autonomous robots is both a relevant and a much-researched topic.
This problem appears in different settings and solutions take on a myriad of different forms. 
In this work, we consider object transportation in the context of space environments, with robots such as the Astrobee~\cite{smith2016astrobee} in the International Space Station, or extra-vehicular robots with tasks such as in-orbit construction~\cite{li2022survey} or deorbeting debris~\cite{jaekel2018design,dubanchet2015modeling}. 
While robots and objects in land, sea, and air environments are subject to friction and gravity, robots and objects in space do not \emph{locally} suffer from this.
When collaborative transportation takes place, this absence should be utilized to its fullest extent. 

\begin{figure}[t!]
    \centering
    \includegraphics[width=0.4\textwidth]{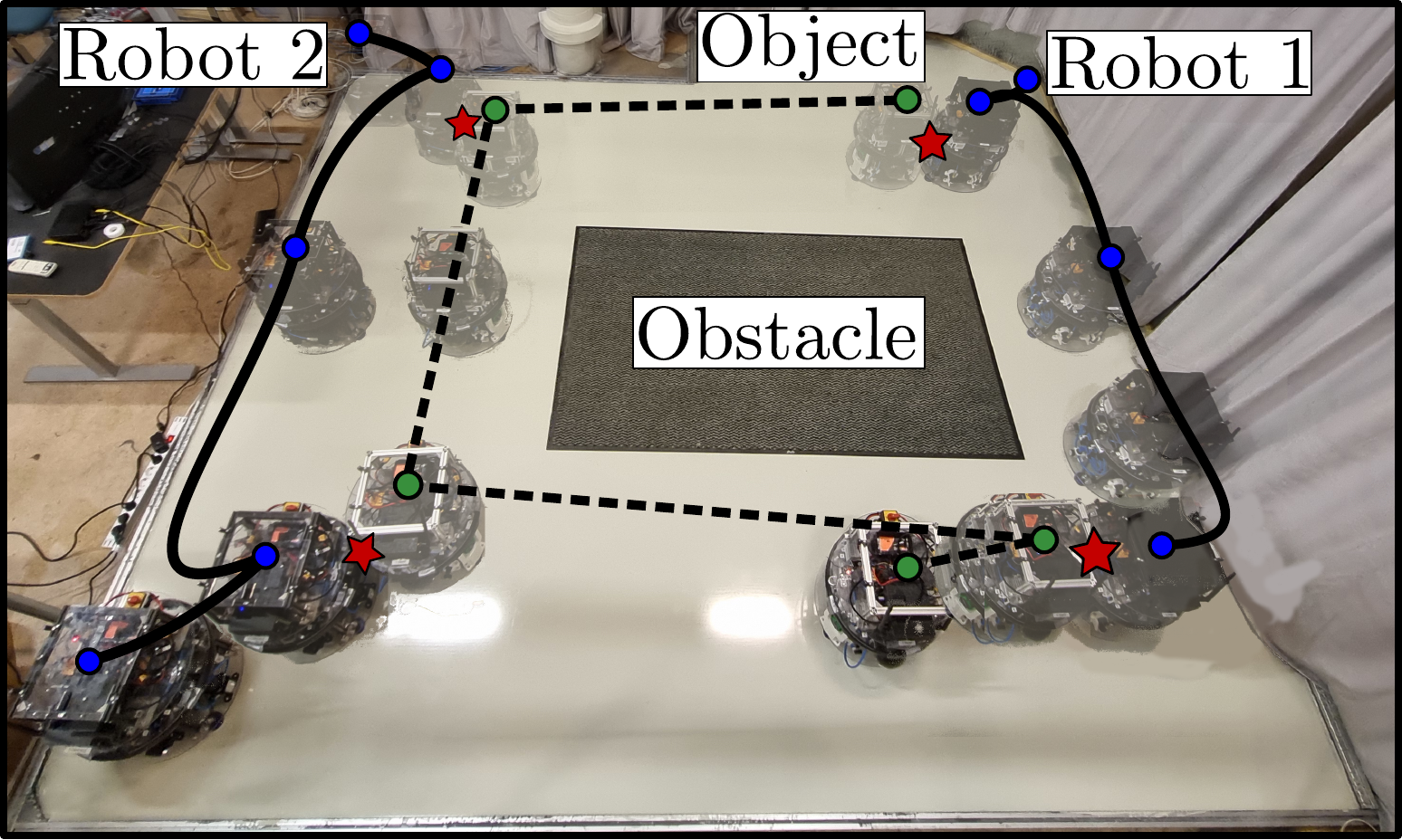}
    \caption{An experimental platform with two controllable freeflyers and a passive object. To transport the object from the top of the figure to the bottom while avoiding the obstacle, Robot 1 impacts it to make it travel towards the top left corner, where Robot 2 impacts it to travel towards the bottom left corner, where it impacts it again to send it towards the right bottom corner where Robot 2 impacts it to reach its desired final destination. 
    Impact times are shown as red stars.
    }
    \label{fig:intro_figure}
\end{figure}

We propose a planning and control approach for collaborative transportation of free-floating objects where impacts are the only robot-object interactions and the only way to change an object's velocity. An example of such scenario in an experimental platform with freeflyers is illustrated in Fig.~\ref{fig:intro_figure}.
The advantage of transportation via impact interaction is its modeling and computational simplicity and its scalability (both computationally and in robot availability). 
Compared to grasping and carrying, this approach also promotes lower energy consumption as well as increased robot availability, since the impacts are minimal, instantaneous, and occasional.
Additionally, it prevents the need of a method of connecting to the object (e.g. via a gripper) as these impacts can occur between objects and robots of any size and shape.

Using the compact formulation of both kinematic impact equations and forward reachability of undisturbed frictionless systems, we are able to model this complex problem as a Mixed-Integer Linear Programming problem (MILP), providing global optimality guarantees even in the presence of complex spatio-temporal goals and constraints.
Although the kinematic impact equations are compact, having them be accurate is a challenge. To that end, we propose an impact-robust planner that explicitly maximizes permissible uncertainty in the impact equation. 

Upon deployment, we consider the interplay between the offline planner, an online replanner and a model predictive controller. 
The offline planner synthesizes a motion plan with simplified point-impact kinematics while the online replanner locally adapts the motion plan around a pre-planned impact to consider updated state information of the approaching object and the physical shapes of the robot and object.
Our planning and control method is shown in a high-fidelity simulator for a range of transportation scenarios and on an experimental freeflyer space platform for 2-robot, 1-object scenarios.
\\\\
Our contributions can be summarized as follows.
\begin{itemize}
    \item We present an impact-aware planner for multi-agent systems with spatio-temporal specifications on controllable freeflyer robots and passive free-floating objects. The planner maximizes either model-agnostic robustness metrics like STL spatial robustness or the uncertainty on the post-impact state of objects.
    \item We then present an impact-aware online planner that relaxes simplifying assumptions on the impact kinematics used in the offline planner and considers updated state information of the free-floating objects.
    \item We present a model predictive control strategy to execute the mission in real-time, which we show in a high-fidelity simulator and on an experimental freeflyer platform.
\end{itemize}

\subsection{Related Work}
There is a significant body of work on collaborative transportation of objects with robots. 
Examples include multi-robot formation transport in dynamics scenarios \cite{alonso2017multi} with parameter and dynamics uncertainty \cite{tagliabue2019robust}.
A distributed control approach has also been proposed for formation transportation in space \cite{farivarnejad2021fully}.
\cite{han2021cooperative} proposes a game-theoretic approach using robots that apply either force or torque to the transported object.
Closest to our approach on planning is \cite{hu2024orbit} which considers a reach-avoid planning problem of an object with rigidly attached robots.

For scenarios where robots are not rigidly attached to objects, a large body of work considers impact dynamics for capturing targets with robotic manipulators \cite{uyama2012experimental,zhang2015pre,huang2016impact}.
Additionally, impact dynamics are considered in docking scenarios \cite{nenchev1999impact,paraskevas2015inertia}. We refer to \cite{flores2014review}, a comprehensive, albeit older, review paper on on-orbit servicing.
In contrast, our work utilizes simpler impact kinematics, due to our online replanning strategy, and explicitly utilizes the free-floating property of objects in space.

Close to our planning and control approach are works on robot air hockey and billiards, where impacts are explicitly used in generating desired behavior of passive objects that float or roll passively. In \cite{jankowski2024airlihockey}, a hierarchical learning-based planner and optimization based controller is utilized to play air hockey against an unknown opponent.
\cite{landry2013heuristic} and \cite{nierhoff2015robotic} consider the game of billiards where a search-based planner is used to pot balls.
In contrast to these games, our scenario allows the robots to replan during execution and to collaborate on achieving a high-level specification. 
\cite{zermane2024minimal} considers minimal pokes to place objects in desired location. The presence of friction makes this problem challenging. We are instead able to utilize kinematic impact models and forward reachability in a globally optimal planner.

In the space of spatio-temporal planning and control, our approach relies on the embedding of STL specifications in mixed-integer optimization problems \cite{raman2014model}. 
The complexity of collaborative specifications ensure that conventional trajectory formulations (via constant time discretization) become intractable. We hence rely on generalizations of these embeddings to piecewise-linear curves \cite{sun2022multi} for objects and B\'ezier curves \cite{verhagen2024temporally} for robots.

%% file: 2_Preliminaries.tex
\section{Preliminaries}
\label{sec:preliminaries}

Let $\mathbb{R}$, $\mathbb{N}$, and $\mathbb{B}$ be the set of real, natural (including zero), and binary numbers respectively. $\mathbb{R}_{\geq 0}$ denotes the set of nonnegative real numbers.

\subsection{System Modelling}
The set of robots is denoted by $\mathcal{R} = \{R_1,...,R_{|R|}\}$ and the set of passive objects by $\mathcal{O} =\{O_1,...,O_{|\mathcal O|}\}$ with $\mathcal{S} = \mathcal{R} \cup \mathcal{O}$ the set of all systems, which are assumed to be rigid bodies. 
Let $x_{S}$ be the state of system $S \in \mathcal{S}$ with $x_{S} = [p_{S},q_{S},\dot{p}_{S},\dot{q}_{S}]$ where $p$ and $q$ indicate position and orientation of the system. 

As we wish to plan for specific impact times, let $\mathit{impact}(S_i,S_j,t) \in \{\top,\bot\}$ be an indicator whether $S_i \in \mathcal{S}$ impacts $S_j \in \mathcal{S}$ at time $t$. Note that  $\mathit{impact}(S_i,S_j,t) \iff \mathit{impact}(S_j,S_i,t)$.
With a slight abuse of notation, let $\mathcal{M}(x_{S_i},x_{S_j},t) : \mathbb{R}^{n} \times \mathbb{R}^{n} \times \mathbb{R}_{\geq 0} \rightarrow \mathbb{R}^{n}$ be an impact model operator, capturing the physics of a collision by taking the states of $S_i$ and $S_j$ at impact time $t$ and responding with the post-impact state of $S_i$ (its first argument).


\subsection{Signal Temporal Logic}
\label{ssec:preliminaries_stl}

Signal Temporal Logic (STL)~\cite{maler2004monitoring} is a formal specification language for dynamical systems. An STL formula is interpreted over real-valued signals qualitatively (the signal satisfies or violates the formula) or quantitatively (the extent to which the signal satisfies or violates the formula).
Time-bounded STL over $n$-dimensional continuous-time signals $\mathbf{x}:\mathbb{R}_{\ge 0} \rightarrow X \subseteq \mathbb{R}^n$ is defined as follows:

\begin{definition}[Time-bounded STL]
\label{def:stl}
    Let $I=[t_1,t_2]$ be a closed bounded time interval,  where $t_1,t_2 \in \mathbb{R}_{\ge 0}, t_1 \leq t_2$.
    Let $\mu:X \rightarrow \mathbb{R}$ be a real-valued function and $p = \mu(x) \geq 0$ a Boolean predicate.
    STL formulas are recursively defined as
    \begin{equation}
        \phi ::= p \mid \neg \phi \mid \phi \land \phi \mid \phi \, \mathcal{U}_I \, \phi,
    \end{equation}
    where $\neg$ and $\land$ are the Boolean negation and conjunction operators and $\mathcal{U}_I$ is the time-bounded Until operator. 
\end{definition}
Formula $\phi_1\mathcal{U}_I \phi_2$ specifies that $\phi_1$ should hold until, within $I$, $\phi_2$ holds. Additional useful temporal operators can be defined as  $\diamondsuit_I \phi = \top \, \mathcal{U}_I \, \phi$ ($\phi$ holds eventually, within $I$), and $\Box_I \phi = \neg \, \diamondsuit_I \, \neg \phi$ ($\phi$ holds at all times within $I$).

\emph{Spatial robustness} is a quantitative way to evaluate satisfaction or violation of a formula by considering the \emph{degree} of satisfaction via the value of the predicate function $\mu$:
\begin{subequations}
\label{eq:stl_space_robustness}
\begin{align}
    \rho_p(t,x) &= \mu(x(t)), \\
    \rho_{\neg \phi}(t,x) &= -\rho_{\phi}(t,x), \\
    \rho_{\phi_1 \land \phi_2}(t,x) &= \min(\rho_{\phi_1}(t,x),\rho_{\phi_2}(t,x)), \\
    \rho_{\phi_1 \mathcal{U}_I \phi_2}(t,x) &= \max_{\tau \in t+I}\big(\min(\rho_{\phi_2}(\tau,x),\min_{s\in[t,\tau]}\rho_{\phi_1}(s,x) ) \big).
\end{align}
\end{subequations}
The reader is referred to~\cite{maler2004monitoring} for a thorough analysis of spatial and also temporal robust semantics of STL.


\subsection{Zonotopes}
\label{ssec:preliminaries_zonotopes}
A zonotope in $\mathbb{R}^n $ is a tuple $Z=<c,G>$, where 
$c \in \mathbb{R}^n$ is the center, and $G \in \mathbb{R}^{n \times k}$ is a set of its $k$ generators. 
We use $V(Z)$ to denote the set of vertices of zonotope $Z$.

Zonotopes enjoy a range of advantageous properties which make them excellent candidates for reachability analysis: they are defined as Minkowski sums of their generators and they are closed under Minkowski sums and linear transformation~\cite{girard2005reachability}.
Additionally, it is straightforward to obtain its interval hull as
\begin{equation}
\label{eq:interval_hull}
    \hat{Z} = <c, diag([\sum_{j=1}^{k}|g_j^1|,...,\sum_{j=1}^{k}|g_j^n|])>,
\end{equation}
which defines a hyperrectangle. 
For trajectory planning, these hyperrectangles can be embedded for reach and avoid behavior using integer variables in mixed-integer programming.

\begin{figure}[t!]
    \centering
    \includegraphics[width=0.45\textwidth]{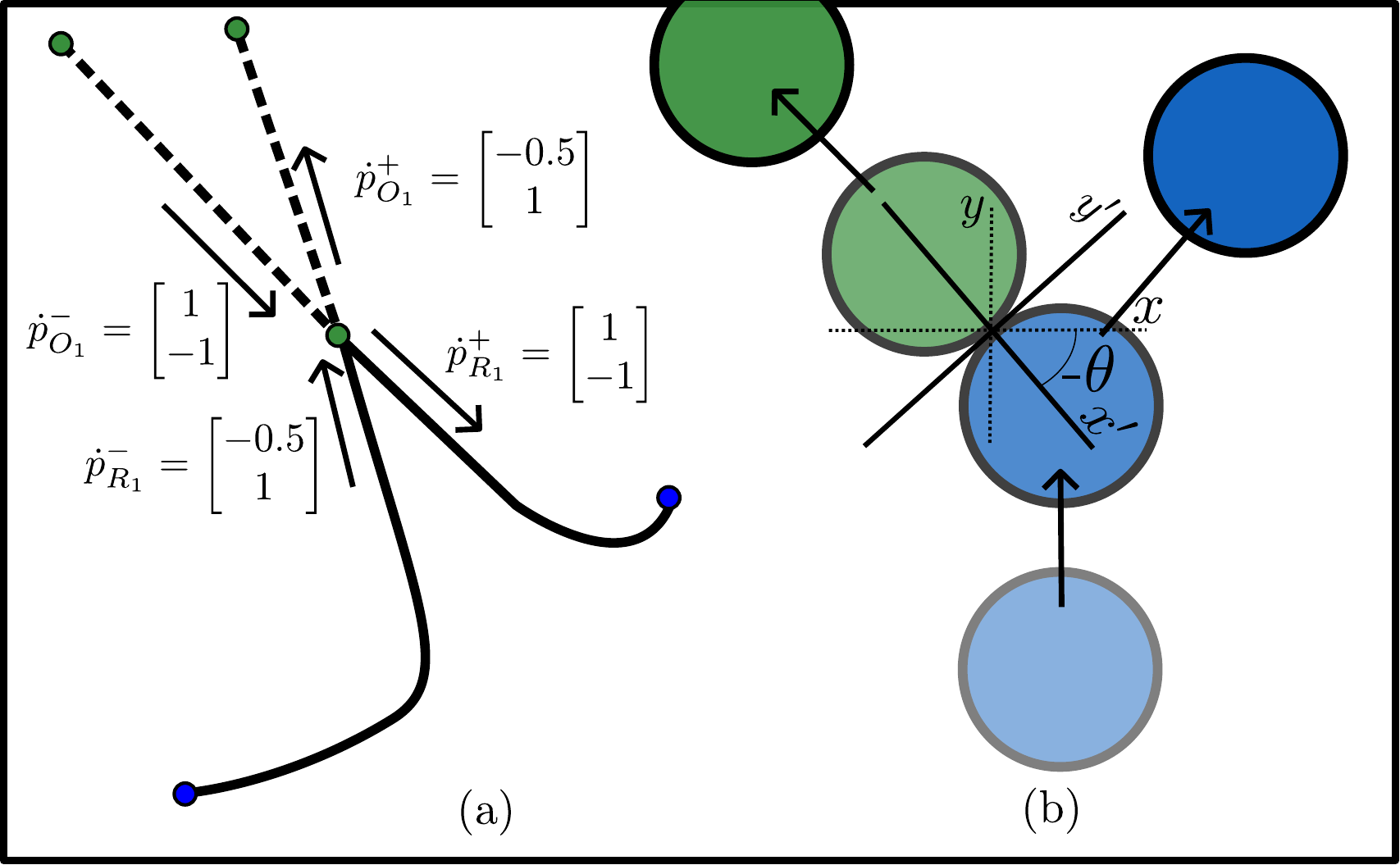}
    
    \caption{(a) Illustration of impact kinematics for point masses (according to Eq.~\eqref{eq:robot_point_impact}). The dashed and solid line indicate a trajectory of the object and robot respectively. Green and blue dots indicated endpoints of B\'ezier curves. The impact occurs after the first B\'ezier curve. (b) Impact kinematics for non-rotating cylindrical objects (according to Eq.~\eqref{eq:robot_cylinder_impact}). The moving blue object impacts a stationary green object. Both object's direction is changed according to the tangential and normal component of the impact on the boundaries of the objects.}
    \label{fig:impact_example}
\end{figure}

\subsection{Impact Kinematics}
\label{ssec:impact_kinematics}

Let $p_{S}(t)$ and $\dot{p}_{S}(t)$ denote the position and velocity of a robot or an object and let $\dot{p}_{S}^{-}(t)$ and $\dot{p}_{S}^{+}(t)$ denote the pre- and post-impact velocity for an impact at time $t$. The general impact model is 
\begin{equation}
    \dot{p}_{S_i}^{+}(t) = \mathcal{M}(\dot{p}_{S_i}^{-}(t),\dot{p}_{S_j}^{-}(t),t).
\end{equation}

For point masses, impact kinematics can be approximated by linear relations between pre- and post-impact velocities.
The post-impact velocities can then be obtained via
\begin{equation}
\label{eq:robot_point_impact}
    \begin{bmatrix} m_{R} & m_{O} \\ 1 & -1 \end{bmatrix} \begin{bmatrix}
        \dot{p}_{R}^{+} \\ \dot{p}_{O}^{+}
    \end{bmatrix} = \begin{bmatrix}
        m_{R}\dot{p}_{R}^{-} + m_{O}\dot{p}_{O}^{-} \\ -e(\dot{p}_{R}^{-} - \dot{p}_{O}^{-})
    \end{bmatrix},
\end{equation}
with its recognizable solution 
\begin{align}
    \dot{p}_{R}^{+} &= \frac{m_{R} - e m_{O}}{m_{R} + m_{O}}\dot{p}_{R}^{-} + \frac{(1+e)m_{O}}{m_{R} + m_{O}}\dot{p}_{O}^{-}, \notag \\
    &= m_1 \dot{p}_{R}^{-} + m_2 \dot{p}_{O}^{-}, \notag \\
    \dot{p}_{O}^{+} &= \frac{(1+e)m_{R}}{m_{R} + m_{O}}\dot{p}_{R}^{-} + \frac{m_{O}-e m_{R}}{m_{R} + m_{O}}\dot{p}_{O}^{-}, \notag \\
    &= m_3 \dot{p}_{R}^{-} + m_4 \dot{p}_{O}^{-}, \notag
\end{align}
with $e \in [0,1]$ being the constant of restitution relating the pre- and post-impact velocities, which is a property of the materials under impact. We use $m_1,m_2,m_3,m_4$ for short-hand notation of the mass fractions.

An example of a robot trajectory (solid line) and an object trajectory (dotted line) of the same mass can be seen in Fig.~\ref{fig:impact_example}(a). For $e=1$ it is a perfectly elastic kinematic impact, i.e. $\dot{p}_{R}^{+} = \dot{p}_{O}^{-}$ and $\dot{p}_{O}^{+} = \dot{p}_{R}^{-}$.


For bodies of a physical size, the pre-impact velocities $\dot{p}_{R}^{-}$ and $\dot{p}_{O}^{-}$ can be split into a tangent and normal component under the assumption of non-rotating bodies or no transfer of rotational momentum during impact.
We can then write the velocity jump as a system of linear equations in the local frame
\begin{align}
\label{eq:robot_cylinder_impact}
    \begin{bmatrix} m_{R} & m_{O} & 0 & 0 \\
    0 & 0 & m_{R} & 0 \\
    1 & -1 & 0 & 0 \\
    0 & 0 & 0 & m_{O} \end{bmatrix} \begin{bmatrix} \dot{p}_{R,x'}^{+} \\ \dot{p}_{O,x'}^{+} \\ \dot{p}_{R,y'}^{+} \\ \dot{p}_{O,y'}^{+} \end{bmatrix} = \begin{bmatrix} m_{R}\dot{p}_{R,x'}^{-} + m_{O}\dot{p}_{O,x'}^{-} \\ m_{R}\dot{p}_{R,y'}^{-} \\ -e (\dot{p}_{R,x'}^{-}-\dot{p}_{O,x'}^{-}) \\ m_{O}\dot{p}_{O,y'}^{-} \end{bmatrix}
\end{align}
where $x'$ and $y'$ denote the tangent and orthogonal axis w.r.t. the plane of impact. 
The post-impact velocities in the global frame can then be obtained via a rotation with $\theta$, the angle between global and local frame,
\begin{equation}
\label{eq:cylinder_impact_rotated}
    \dot{p}_{S}^{+} = R^{-1}(\theta)\begin{bmatrix} \dot{p}_{S,x'}^{+} \\ \dot{p}_{S,y'}^{+} \end{bmatrix}.
\end{equation}
A schematic example of point-mass and cylindrical body impacts with its global and local axis is shown in Fig.~\ref{fig:impact_example}(b). 




\begin{figure}
    \centering
    \includegraphics[width=0.48\textwidth]{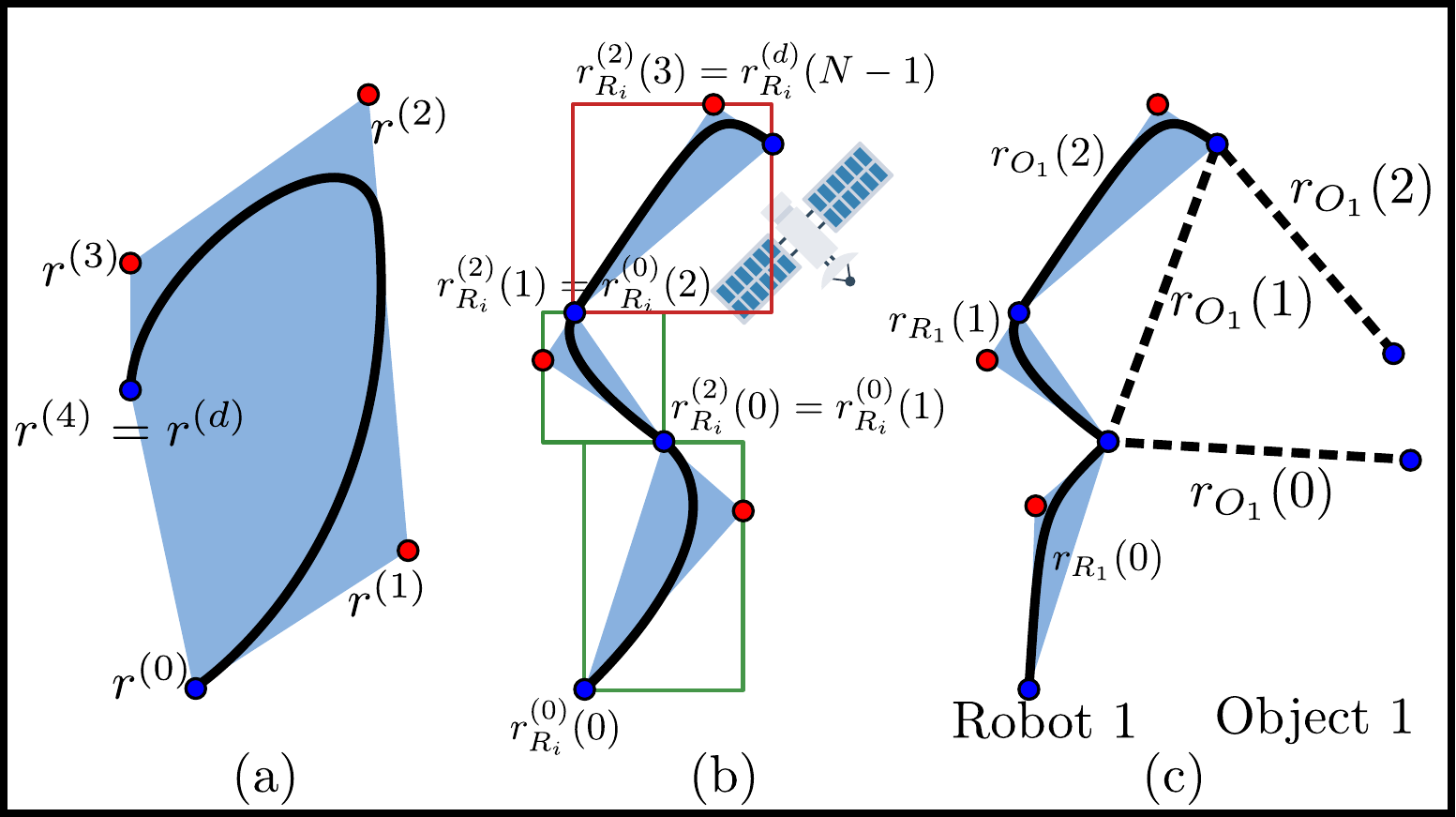}
    \caption{(a) a B\'ezier curve with its control points (red dots) and convex hull (shaded blue), (b) a trajectory of robot $R_i$ with three B\'ezier curves. Note the bounding-box used for collision avoidance, (c) the B\'ezier trajectory formulation for objects (first degree B\'ezier curves) and robots (higher-degree B\'ezier curves and continuity conditions). $r_{R_1}(0)$ and $r_{R_1}(1)$ are the pre- and post-impact curves of impact $1$. $r_{R_1}(1)$ and $r_{R_1}(2)$ are the pre- and post-impact curves of impact $2$.}
    \label{fig:parametrization}
    \vspace{-0.5cm}
\end{figure}

\subsection{B\'ezier Curves}
\label{ssec:preliminaries_bezier_curves}
B\'ezier curves are a popular tool in trajectory planning of autonomous systems \cite{qian2016motion,marcucci2023motion,pant2018fly}. 
A B\'ezier curve is represented by a polynomial equation and parameterized by a finite number of control points, its decision variables.
Namely, a B\'ezier curve $r$ of degree $d$ is constructed and evaluated using the summation of its $d$ Bernstein polynomials multiplied with their respective $d$ control points according to
\begin{equation}
\label{eq:bezier_evaluating}
    r(s) := \sum_{b=0}^{d}\begin{pmatrix}d \\ b\end{pmatrix}(1-s)^{d-b}s^b \cdot r^{(b)}, \quad s\in[0,1],
\end{equation}
where $s$ is the phasing parameter, $\begin{pmatrix}d \\ b\end{pmatrix}(1-s)^{d-b}s^b$ is the $b$'th Bernstein polynomial, and $r^{(b)}$ is the $b$'th control point. 
An example of a B\'ezier curve with its control points is shown in Fig. \ref{fig:parametrization}.(a).
Although B\'ezier curves are nonlinear in nature, the following properties allow us to reason over their convex over-approximation:
\begin{enumerate}
    \item \textit{Convex hull:} the curve $r(s)$ is entirely contained within the convex hull generated by its control points $r^{(b)}$, $r(s) \in \mathrm{hull}_{b=0}^{d}(r^{(b)}) \quad \forall s \in [0,1]$. A more conservative but closed-form overapproximation is the \emph{bounding-box property} obtained by using the interval hull, $r(s) \in [\min_{b=0}^{d}(r^{(b)}),\max_{b=0}^{d}(r^{(b)})]$ which we use to embed collision avoidance.
    \item \textit{Endpoint values:} the curve $r(s)$ starts at the first control point $r^{(0)}$ as $s=0$ and ends at the last control point $r^{(d)}$ as $s=1$.
    \item \textit{Derivatives:} the derivative $\dot{r}(s) = \frac{dr(s)}{ds}$ is a linear combination of the control points of $r(s)$ and is again a B\'ezier curve of degree $d-1$ with control points $\dot{r}^{(b)} = d\cdot(r^{(b+1)} - r^{(b)})$ for $b\in\mathbb{N}_{[0,...,d-1]}$.
\end{enumerate}
Subsequently, the derivatives of the start- and end-point, $\dot{r}^{(0)}$ and $\dot{r}^{(d-1)}$, are linear combinations of the control points of $r(s)$ and the interval hull of $\dot{r}$ is formed as a linear combination of its control points.
This means that smooth, kinematically feasible, and collision-free trajectories can be constructed through Mixed-Integer Linear Programming by returning a finite set of control points of a B\'ezier curve as illustrated in Fig.~\ref{fig:parametrization}.(b).

%% file: 3_Problem.tex
\section{Problem Formulation and Approach}
\label{sec:problem_formulation}

We wish to solve the problem of collaborative transportation of passive objects in a weightless environments by a team of robots. 
The transportation tasks as well as the goals for the system, obstacle avoidance, and possible additional constraints are specified via STL.
The robots interact with the objects via an instantaneous impact. 
Formally, let $R_i \in \mathcal{R}$ and $O_i \in \mathcal{O}$ be controllable robots and uncontrollable objects. 
We aim to find a controller $u_{R_i}, \forall R_i \in \mathcal{R}$, such that the robot and the object trajectories are maximally robust. Furthermore, we require STL satisfaction, kinematic constraints and smoothness, and collision avoidance. 

\subsection{Problem formulation}

We formulate the problem of collaborative transportation via impact interactions as
\begin{subequations}
\label{eq:opt_prob}
\begin{align}
    &\min_{x_{R_i},u_{R_i},x_{O_j}} J(\mathbf{x}_{\mathcal R},\mathbf{u}_{\mathcal R},\mathbf{x}_{\mathcal O}), \tag{\ref*{eq:opt_prob}} \\
    \textrm{s.t.} \hquad & [\mathbf{x}_{\mathcal R},\mathbf{x}_{\mathcal O}] \models \phi, \label{eq:opt_phi} \\
    & \dot{\mathbf{x}}_{\mathcal R} \in \mathcal{V}_{\mathcal R}, \dot{\mathbf{x}}_{\mathcal O} \in \mathcal{V}_{\mathcal O}, \label{eq:opt_vel}\\
    & \dot{x}_{R_i} =   \begin{cases}
                            \begin{bmatrix} \mathcal{M}(x_{R_i}(t),x_{O_j}(t)) & 0 \end{bmatrix}^T \textrm{if } \mathit{impact}(R_i,O_j,t) \\
                            \begin{bmatrix} \dot{p}_{R_i} & \dot{\theta}_{R_i} & u_{R_i} \end{bmatrix}^T \textrm{otherwise}
                        \end{cases} \label{eq:opt_control_R}\\
    & \dot{x}_{O_i} =   \begin{cases}
                            \begin{bmatrix} \mathcal{M}(x_{O_i}(t),x_{R_j}(t)) & 0 \end{bmatrix}^T\textrm{if } \mathit{impact}(O_i,R_j,t)\\
                            \begin{bmatrix} \dot{p}_{O_i} & \dot{\theta}_{O_i} & 0 \end{bmatrix}^T\textrm{otherwise}
                        \end{cases} \label{eq:opt_control_O}\\
    & \forall i,j \in |S|, i \neq j, \label{eq:opt_col}, \forall t\in[t_0,t_f]:  \neg \textit{impact}(S_i,S_j,t) \iff  \notag \\
    & ||p_{S_i}(t) - p_{S_j}(t)||_2 \geq \mathrm{rad}_{S_i} + \mathrm{rad}_{S_j},
\end{align}
\end{subequations}
where $x_{S_i} = [p,\theta,\dot{p},\dot{\theta}]^T \in \mathbb{R}^{2n}$ is the state of a system $S_i \in \{O,R\}$, and $\mathbf{x}_{\mathcal{R}}$ and $\mathbf{x}_{\mathcal{O}}$ denote the conjoined state of all robots and objects respectively. 
$\mathcal{V}_{\mathcal R}$ and $\mathcal{V}_{\mathcal O}$ denote the permissible velocity set of the robots and objects respectively and $\mathrm{rad}_{S_i}$ the bounding radius of $S_i$.
$u_{R_i} \in \mathbb{R}^n$ denotes the control input of $R_i$.
We assume workspace constraints and obstacle avoidance are captured in $\phi$ (e.g. $\Box_{[t_0,t_f]}(x_{S_i} \notin Obs)$).

Eq.~\eqref{eq:opt_phi} ensures that the trajectories of the robots and objects satisfy the high-level specification. 
Eq.~\eqref{eq:opt_vel} are the velocity constraints.
Eq. ~\eqref{eq:opt_control_R} and~\eqref{eq:opt_control_O} ensure the conditional smoothness and kinematics of the robot and object respectively, dependent on the presence of an impact between a robot and an object. Eq.~\eqref{eq:opt_col} ensures collision avoidance between robots and objects, except at moments when impact is desired.

\begin{example}
    Consider a 2-robot, 1-object scenario; $\mathcal{R}=\{R_1,R_2\}$, $\mathcal{O}=\{O_1\}$, $\mathcal{S}=\{R_1,R_2,O_1\}$. 
    We use a corridor transportation scenario as a running example, as shown in Fig.~\ref{fig:corridor_space_robust}(a). 
    The robots in the bottom and top of the figure, both depicted as a black dots, aim to transport the object depicted with white details to the green square area in the top before time runs out, corresponding to an STL specification $\phi_{O_1}= \diamondsuit_{[t_0,t_f]}(x_{O_1} \in \mathcal{X}_f)$. 
    Additionally, the robots and the object should avoid collision with corridor walls, corresponding to STL specification $\phi_{S_i} = \Box_{[t_0,t_f]} (x_{S_i} \in \mathcal{W}_{\textrm{free}})$, $\forall S_i \in \mathcal{S}$. 
    To transport the object from a static initial position to a static final position, the plan requires at least two impacts; a \emph{throw} and a \emph{catch}.
\end{example}

\subsection{Approach}

We take a hierarchical approach to address the problem. We define an offline planner that, using simplified kinematic impact models, can efficiently solve the multi-agent planning problem with STL specifications via a B\'ezier curve trajectory formulation. A part of this plan is when and where robots impact objects.  An online replanner that corrects the plans of the two B\'ezier curves leading up to and following an impact. The replanner considers a more complex impact model and the updated states of the robots and objects during execution. Finally, a Model Predictive Controller (MPC) that realizes tracking of the planned B\'ezier curves and ensures the pre-impact conditions to enable the desired post-impact behavior of the robot.

We describe the offline planner in Sec~\ref{sec:method} with STL spatial robustness as the optimization function in Sec.~\ref{ssec:method_best_case} and an impact-robust optimization metric in Sec.~\ref{ssec:method_worst_case}. In Sec.~\ref{sec:implementation} we describe how this planning problem is implemented as a mixed-integer optimization problem.
The online replanner is detailed in Sec.~\ref{sec:replanner} and the impact-aware MPC in Sec.~\ref{sec:implementation_impact_aware_mpc}.



%% file: 4_Method.tex
\section{Offline Planner}
\label{sec:method}

In the offline planner, we use a simplistic point-mass kinematic impact model which ensures that all constraints regarding impacts can be encoded linearly. 
Additionally, for simplicity, we do not consider transfer of rotational momentum between impacts. Instead, we abstract the planning aspect of Eq.~\eqref{eq:opt_prob} to a kinematically constrained geometric optimization problem. 
In summary, we make two additional assumptions:
\begin{assumption}[Planner Impact Modeling]
\label{ass:no_radius_impact}
    For the modeling of the impact kinematics in the planner, the radii $r_{R_i}$ and $r_{O_i}$ of robots $R_i \in \mathcal{R}$ and objects $O_i \in \mathcal{O}$ respectively are assumed zero, and hence $R_i$ and $O_i$ are considered points.
\end{assumption}


\begin{assumption}[Planner State Modeling]
    The state $x_{S_i}$ is reduced to $x_{S_i}=[p_{S_i},\dot{p}_{S_i}]$ for the planner, meaning we ignore rotation of the robots and objects.
\end{assumption}

Relaxing these assumptions in the offline planner would introducing the nonlinear impact model of Eq.~\eqref{eq:robot_cylinder_impact} which, for planning, introduces non-convexities, making it difficult to obtain global optimal solutions w.r.t. the specifications $\phi_{S_i}$. 
Inaccuracies introduced by these assumptions will be treated by the replanner and the controller.



For the robots, we consider integrator dynamics with unconstrained (but attenuated, see Sec. \ref{sec:implementation}) acceleration inputs and the possibility of discrete jumps in velocities due to an impact with an object. As such, we define the impact model operator 
for the offline planner as
\begin{equation}
    \label{eq:M_planner}
    \mathcal{M}_{\textrm{planner}}(x_{S_i},x_{S_j}) = \frac{m_{S_i}-e m_{S_j}}{m_{S_i}+m_{S_j}}\dot{p}^{-}_{S_i} + \frac{(1+e)m_{S_j}}{m_{S_i}+m_{S_j}}\dot{p}^{-}_{S_j},
\end{equation}
according to the point-mass impact equations in Eq.~\eqref{eq:robot_point_impact}.




\subsection{Trajectory Formulation}
\label{ssec:method_trajectory}
Finding a plan to satisfy Eq.~\eqref{eq:opt_prob} inadvertently includes a decision on \emph{if}, \emph{when}, and with \emph{whom} to impact according to Eqs.~\eqref{eq:opt_control_R}~\eqref{eq:opt_control_O} and~\eqref{eq:opt_col}.
Fine-grained constant-time discretization of the problem would lead to a large number of possible impact points to be considered; later on in the MILP formulation of the planner this would project onto a large number of binary variables.
In order to reduce the computational complexity, we parametrize the trajectory of robots and objects with B\'ezier curves.
We define a \emph{curvature} B\'ezier curve $r(s)$ and a \emph{temporal} B\'ezier curve $t := h(s)$.
We then couple these curves to parameterize the physical trajectory over time.
We base our multi-agent B\'ezier formulation on the work in \cite{marcucci2023motion} which has been explored for high-level specifications with STL in \cite{verhagen2024temporally}. 
In these works, the authors express a trajectory curve $p$ via the use of $r(s)$ and $h(s)$ as
\begin{align}
    r(s) &:= p(h(s)), \notag \\
    \dot{r}(s) &:= \dot{p}(h(s))\dot{h}(s), \label{eq:bezier_formulation}
\end{align}
where $p$ is the physical trajectory of interest, parameterized by the two types of B\'ezier curves. 
The main advantage of this formulation is that 
we may generate trajectories spanning long distances and durations with a small number of B\'ezier curves and hence a small number of variables. The computational speedup in the context of STL was shown in~\cite{verhagen2024temporally}.

For objects, $r(s)$ and $h(s)$ are limited to 1st degree B\'eziers (straight line curves) as only impacts with robots may instantaneously change their velocity. 
An indicative example of the parametrization, in addition to requirements on collision avoidance is shown in Fig. \ref{fig:parametrization}(c).
Note that impacts may only occur at endpoints of B\'ezier curves and that an impact subsequently assigns a \emph{pre-impact} label to that B\'ezier curve and a \emph{post-impact} label to the proceeding curve.


\begin{figure*}[t]
    \centering
    \includegraphics[width=0.98\textwidth]{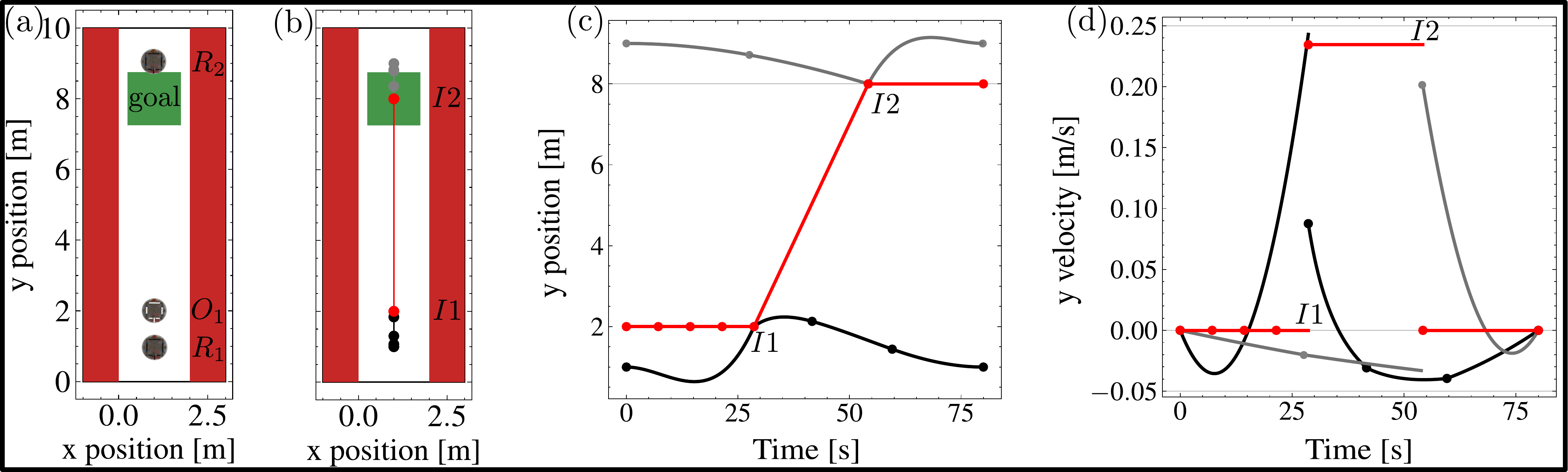}
    \caption{An example corridor travel scenario with (a) the problem setup with 2 robots ($R_1$ and $R_2$) and one object ($O_1$). The objective is to transport $O_1$ to \emph{goal} while all systems should avoid the red obstacles. (b) the spatially-robust plan from the offline planner in \ref{ssec:method_best_case} with two impacts denoted by $I_1$ and $I_2$. (c) the vertical position over time. (d) the vertical velocity over time. Notice that the velocity (and therefore position) of $O_1$ can only be changed via impacts with $R_1$ or $R_2$. The change in velocity is instantaneous, according to the impact kinematics in Eq.~\eqref{eq:robot_point_impact}.}
    \label{fig:corridor_space_robust}
\end{figure*}

\subsection{Spatially Robust Offline Planner}
\label{ssec:method_best_case}

We will discuss two different options of integrating the impact kinematics into the planner: without and with uncertainty in the impact kinematics. This section focuses on the former and the following section on the latter.

We embed the deterministic impact condition using Eq.~\eqref{eq:robot_point_impact} 
\begin{align}
    \begin{bmatrix} m_{R} & m_{O} \\ 1 & -1 \end{bmatrix} \begin{bmatrix}
        \dot{p}_{R}^{+} \\ \dot{p}_{O}^{+}
    \end{bmatrix} = \begin{bmatrix}
        m_{R}\dot{p}_{R}^{-} + m_{O}\dot{p}_{O}^{-} \\ -e(\dot{p}_{R}^{-} - \dot{p}_{O}^{-})
    \end{bmatrix} \iff p_{R}^{-} = p_{O}^{-},
\end{align}
for each curve of the interacting robot and object. 
Additionally, if no impact occurs, the B\'ezier curves of robots and objects should be smooth via
\begin{align}
    \dot{x}_{R}^{+}(k+1) = \dot{x}_{R}^{-}(k) &\iff p_{R}^{-} \neq p_{O_i}^{-}, \forall O_i \in \mathcal{O}, \\
    \dot{x}_{O}^{+}(k+1) = \dot{x}_{O}^{-}(k) &\iff p_{R_i}^{-} \neq p_{O}^{-}, \forall R_i \in \mathcal{R},
\end{align}

Although we do not explicitly consider any uncertainty in the robot dynamics or impact kinematics, a common choice in trajectory planning with spatio-temporal specifications such as STL, is to maximize its spatial robustness. As the B\'ezier curve is a generalization of a piece-wise linear trajectory, we can rely on \cite{sun2022multi} or \cite{verhagen2024temporally} to embed STL spatial robustness
according to the definitions in Eq.~\eqref{eq:stl_space_robustness}. 
The spatially robust planner problem is formulated as 
\begin{subequations}
\label{eq:opt_model_free}
\begin{align}
    &\max_{r_{R_i},h_{R_i},r_{O_j},h_{O_j}} J(r_{R_i},h_{R_i},r_{O_j},h_{O_j}) = \rho_{\phi}(\mathbf{x}_{\mathcal R},\mathbf{x}_{\mathcal O}), \tag{\ref*{eq:opt_model_free}} \label{eq:bc_planner_cost} \\
    \textrm{s.t.} \notag \\ & \dot{x}_{R_i} =   \begin{cases}
                            \begin{bmatrix} \mathcal{M}_{\textrm{planner}}(x_{R_i}(t),x_{O_j}(t)) & 0 \end{bmatrix}^T \\ \hspace{4cm} \textrm{if }\textit{impact}(x_{R_i},x_{O_j},t)\\
                            \begin{bmatrix} \dot{p}_{R_i} & u_{R_i} \end{bmatrix}^T   \textrm{otherwise}
                        \end{cases} \label{eq:bc_planner_control_R}\\
    & \dot{x}_{O_j} =   \begin{cases}
                            \begin{bmatrix} \mathcal{M}_{\textrm{planner}}(x_{O_j}(t),x_{R_i}(t)) & 0 \end{bmatrix}^T \\ \hspace{4cm} \textrm{if }\textit{impact}(x_{O_j},x_{R_i},t)\\
                            \begin{bmatrix} \dot{p}_{O_j} & 0 \end{bmatrix}^T \textrm{otherwise} \\
                        \end{cases} \label{eq:bc_planner_control_O}\\
    & \eqref{eq:opt_phi}, \eqref{eq:opt_vel}, \eqref{eq:opt_col}, \forall R_i \in \mathcal{R}, O_i \in \mathcal{O},\label{eq:bc_planner_others}
\end{align}
\end{subequations}
where we reiterate that for the planner, $x_{S_i} = [p_{S_i},\dot{p}_{S_i}]^T$ and $p_{S_i}$ and $\dot{p}_{S_i}$ can be constructed from the B\'ezier curves $r_{S_i}$ and $h_{S_i}$ according to Eq.~\eqref{eq:bezier_formulation}.

The solution of Eq.~\eqref{eq:bc_planner_cost} ensure that $\mathcal R$ and $\mathcal O$ satisfy $\phi$ with maximal spatial robustness, adhering to velocity constraints, avoiding collisions between objects and robots and resolving the impact times and positions. 

\begin{example}
    Consider again the corridor scenario in Fig.~\ref{fig:corridor_space_robust}(a).
    The initial and final static positions for $R_1$ and $R_2$ are $[1,0]$ and $[8,0]$ respectively. 
    As we maximize the spatial robustness of $\phi = \bigwedge_{S_i} \phi_{S_i}$, $O_1$ is pushed by $R_1$ to move towards the goal, maximizing the distance the walls and the goal's boundaries ($\rho_{\phi}=0.4m$). $R_2$ catches $O_1$ and brings it to a standstill.
    Fig.~\ref{fig:corridor_space_robust}(b), (c) and (d) show how smooth B\'ezier curves and piecewise linear curves construct the robot and object trajectories. At collision, the velocity of the robot and object is changed instantaneously.
\end{example}

\subsection{Impact-Robust Offline Planner}
\label{ssec:method_worst_case}
The spatially robust planner can accommodate uncertainties arising in the execution of the planned trajectory in terms of STL spatial robustness. 
This model-free robustness metric is not explicitly related to the most significant contributor of uncertainty; the impact kinematics. 
To address  inaccurate tracking of the pre-impact trajectory (in position or velocity), the kinematic model assumptions, or inaccurate model constants, we introduce the impact-robustness term $\delta$, 
which captures the permissible uncertainty in the post-impact velocity of the objects.

First, consider uncertainty in the initial position of an object $O$, represented as an initial set $X_0$ contained by the zonotope $Z_{O,X_0} \subset \mathbb{R}^n$.
By ensuring kinematically feasible robot trajectories leading up to the desired impact with $O$ upon execution 
we ensure that the robot can generate the desired post-impact state on $O$ for all possible states $x_{O} \in Z_{O,X_0}$. 
If we consider an interval-hull overapproximation $\hat{Z}_{O,X_0}$ (via Eq.~\eqref{eq:interval_hull}), this amounts to doing so only for extremal vertices according to the following Lemma.

\begin{lemma}
\label{lemma:v_in_X}
    Let $\hat{Z}$ be a interval hull zonotope of zonotope $Z$. If for all $v \in V(\hat{Z})$ there exists a trajectory $x_{R,v}(t_0:t_f)$ that intersects $v$ at $t_f$, then for all $\forall x \in Z$ there exist a trajectory that interests $x$ at $t_f$.
\end{lemma}

Considering an impact on all extremal vertices $v \in V(\hat{Z})$ requires the creation of $2^n$ B\'ezier trajectories, as this is the number of vertices of an $n$-dimensional interval hull.

Let $p_v$ denote the position at an outer vertex $v \in V(\hat{Z})$ of the interval hull zonotope of $Z$, we then obtain the kinematic impact equation $\forall v \in V(\hat{Z})$
\begin{equation}
\label{eq:robust_velocity_impact_robot}
    \dot{p}_{v,R}^{+} = \begin{cases}
        m_1 \dot{p}_{v,R}^{-} + m_2 \dot{p}_{v,O}^{-} \pm \delta, & \iff p_{v,R}^{-} = p_{v,O}^{-} \\
        \dot{p}_{v,R}^{-}, & \textrm{otherwise},
    \end{cases} 
\end{equation}
\begin{equation}
\label{eq:robust_velocity_impact_object}
    \dot{p}_{v,O}^{+} = \begin{cases}
        m_3 \dot{p}_{v,R}^{-} + m_4 \dot{p}_{v,O}^{-} \pm \delta, & \iff p_{v,R}^{-} = p_{v,O}^{-} \\
        \dot{p}_{v,O}^{-}, & \textrm{otherwise}.
    \end{cases} 
\end{equation}
The uncertainty term $\delta$ represents a deviation from the nominal post-impact velocity of the object.
The remainder of this section is devoted to formulating the planning problem so that the optimization metric in Eq. \eqref{eq:opt_prob} is to maximize $\delta$, i.e. the optimal plan is the most robust with respect to post-impact velocity uncertainty.

\medskip

\begin{figure*}[t]
    \centering
    \includegraphics[width=0.98\textwidth]{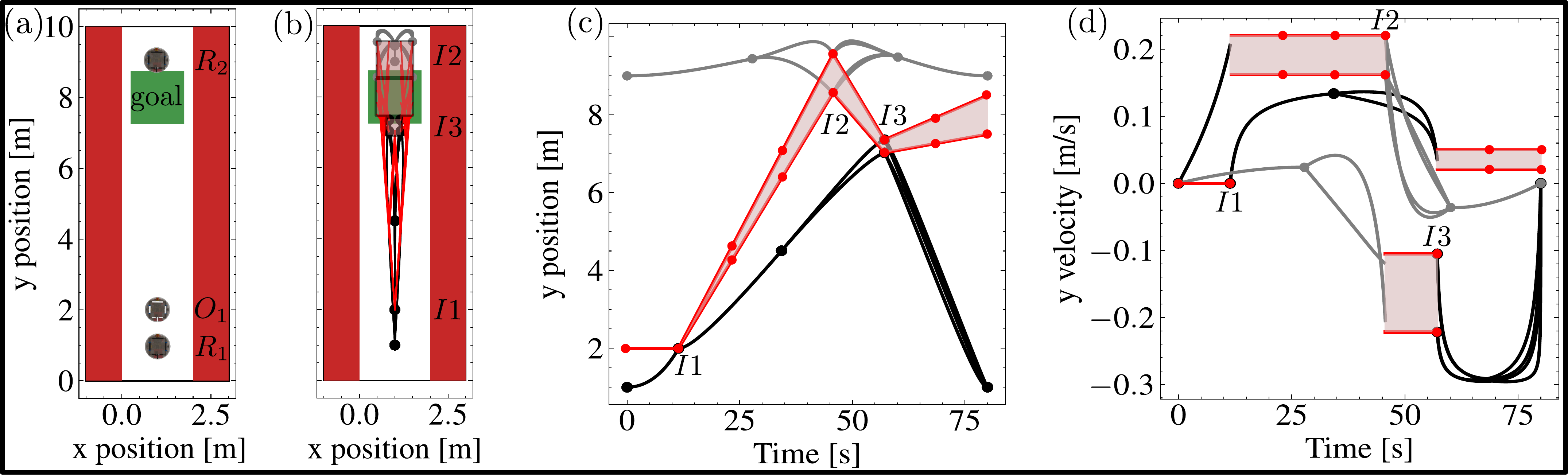}
    \caption{An example corridor travel scenario with (a) the problem setup with 2 robots ($R_1$ and $R_2$) and one object ($O_1$). The objective is to transport $O_1$ to \emph{goal} while all systems should avoid the red obstacles. (b) the impact-robust plan from the offline planner in \ref{ssec:method_worst_case}. Three planned impacts are indicated by $I_1$, $I_2$ and $I_3$. Instead of a deterministic state for $O_1$ as in Fig.~\ref{fig:corridor_space_robust}, the post-impact states are now represented by zonotopes and the robot trajectories consider the traversal to all extremal vertices. (c) the vertical position over time. Impact $I_2$ at $t\approx 45s$ creates a decreasing funnel indicating that the overall uncertainty in the system can be reduced by additional impacts (as all $v \in \hat{Z}_{O}$ have a unique impact). (d) the vertical velocity over time. Notice that the velocity (and therefore position) of $O_1$ can only be changed via impacts with $R_1$ or $R_2$ and that any impact between $O_1$ and $R_i$ introduces the propagation of a funnel due to the uncertain post-impact term $\delta$. 
    Note the constant time duration of of the zonotope segments of the object in (c) and (d).}
    \label{fig:corridor_impact_robust}
\end{figure*}

For the uncertainty propagation of the uncertain set $\hat{Z}_{O,X_0}$ after impact, we use zonotope propagation on each vertex $v \in \hat{Z}_{O,X_0}$ with the uncertainty $\delta$ in the post-impact velocity
\begin{equation}
\label{eq:robust_impact_object}
    Z^{+}_{v,O} = \begin{cases}
        <\begin{bmatrix} p_{v,O} \\ \dot{p}_{v,O} \end{bmatrix}, \textrm{diag}(0,0,\delta,\delta)>, & \iff p_{v,R}^{-} = p_{v,O}^{-} \\
        <\begin{bmatrix} p_{v,O} \\ \dot{p}_{v,O} \end{bmatrix}, \textrm{zeros}_{n\times k}>, & \textrm{otherwise},
    \end{cases} 
\end{equation}
which ensures that uncertainty in the post-impact velocity translates to velocity and spatial uncertainty in the state propagation
\begin{equation*}
    Z^{-}_{v,O} = e^{A\Delta t} Z^{+}_{v,O} = <e^{A\Delta t}c, e^{A\Delta t}G>,
\end{equation*}
where $c$ and $G$ are from Eq.~\eqref{eq:robust_impact_object}, dependent on the impact condition.
Since the system under consideration is a double integrator without friction, as it is a floating object in space, we can define the matrix exponential as $e^{A\Delta t} = (I+A\Delta t)$.
\begin{equation}
\label{eq:IAT_forward_reach}
    Z^{-}_{v,O} = (I+A\Delta t) Z^{+}_{v,O} = <(I+A\Delta t)c, (I+A\Delta t)G>.
\end{equation}
Note that in the trajectory formulation in Eq. \eqref{eq:bezier_formulation}, $\Delta t$ is the duration of the temporal B\'ezier curve, $h(s)$, which is a continuous variable. 
This means that as $\Delta t$, $p_{v,O}$, $\dot{p}_{v,O}$, and $\delta$ are continuous variables, Eq.~\eqref{eq:IAT_forward_reach} introduces bilinear constraints. 
We remedy this by choosing a constant time-discretization for the objects.

We continue this forward reachability process for all trajectory curves by obtaining the pre-impact interval hull $\hat{Z}^{-}_{O} \supseteq Z^{-}_{v,O}, \forall v \in \hat{Z}_{O,X_0}$ using Eq.~\eqref{eq:interval_hull}.
Then, the forward reachability procedure repeats itself for all impacts and non-impacts according to Eqs.~\eqref{eq:robust_velocity_impact_object} and~\eqref{eq:robust_impact_object}. An example of this procedure is schematically shown in Fig.~\ref{fig:corridor_impact_robust}.

The impact-robust planner problem is then formulated as 
\begin{subequations}
\label{eq:opt_model_based}
\begin{align}
    &\max_{r_{R_i},h_{R_i},r_{O_i}} J(r_{R_i},h_{R_i},r_{O_i}) =  \delta, \tag{\ref*{eq:opt_model_based}} \\
    \textrm{s.t.} \notag \\& \dot{x}_{v,R_i} =   \begin{cases}
                            \begin{bmatrix} \mathcal{M}_{\textrm{planner}}(x_{v,R_i}(t),v \in \hat{Z}_{O_j}(t)) & 0 \end{bmatrix}^T \\ 
                            \begin{bmatrix} \dot{p}_{v,R_i} & u_{v,R_i} \end{bmatrix}^T \\
                        \end{cases} \label{eq:ir_planner_control_R}\\
    & \dot{x}_{O_i} =   \begin{cases}
                            \begin{bmatrix} \mathcal{M}_{\textrm{planner}}(v \in \hat{Z}_{O_j}(t),x_{v,R_j}(t)) & 0 \end{bmatrix}^T \\
                            \begin{bmatrix} v_{\dot{p}} \in \hat{Z}_{O_j} & 0 \end{bmatrix}^T \\
                        \end{cases} \label{eq:ir_planner_control_O}\\
    & \eqref{eq:opt_phi}, \eqref{eq:opt_vel}, \eqref{eq:opt_col}, \label{eq:bc_planner_others}
\end{align}
\end{subequations}
where $v_{\dot{p}}$ refers to the velocity states of the vertex $v \in \hat{Z}_{O_j}$. 
Notice that the impact-robust planning problem in Eq.~\eqref{eq:opt_model_based} does not optimize for the duration of the object curve ($h_{O_i}$) but does face the challenge of increased overall curves compared to the spatially robust planning problem in Eq.~\eqref{eq:bc_planner_cost}.

\begin{example}
    Consider again the corridor transportation scenario. 
    The initial state of the object is now defined as a bounded set, $Z_{O_1,X_0} = \hat{Z}_{O_1,X_0}$.
    As we maximize the impact-robustness, $\delta$, we consider how, from each $v \in \hat{Z}_{O_1,X_0}$, uncertainty in the post-impact velocity translates to spatial and velocity uncertainty in the next pre-impact zonotope. 
    The increase from $\hat{Z}_{O_1,X_0}$ to $\hat{Z}_{O_1}^{-}$ is shown in Fig.~\ref{fig:corridor_impact_robust}. 
    This also highlights that we consider $2^n$ trajectories of the robots for the feasibility $\forall x \in \hat{Z}_{O_1,X_0}$ according to Lemma.~\ref{lemma:v_in_X}.
    A second impact with $R_1$, while also introducing a post-impact uncertainty, allows redirecting the object. 
    This kind of behavior is implicitly embedded via the impact-robustness metric $\delta$ and not apparent when maximizing a model-free robustness metric.
\end{example}


%% file: 5_Implementation.tex
\section{MILP encoding of the offline planner}
\label{sec:implementation}

Given the planner from Sec.~\ref{sec:method}, we present its MILP encoding. 
We detail how the constraints in Eq.~\eqref{eq:opt_phi}-\eqref{eq:opt_col} are embedded in the offline MILP planner and how accelerations of the robots can be attenuated using quadratic cost terms. 

\subsubsection{Specification $\phi$}
For the satisfaction of $\phi$ for $\mathbf{x}_{R}$ and $\mathbf{x}_{O}$ in Eq.~\eqref{eq:opt_phi}, we refer for details to \cite{sun2022multi} and \cite{verhagen2024temporally}. The key-takeaway is that each Bézier curve is assigned a robustness value w.r.t. space- or time-robustness, $\rho_{p}(x_{S,i})$ or $\theta_{p}(x_{S,i})$. Using the recursive definitions in Eq.~\eqref{eq:stl_space_robustness}, the robustness of trajectory $x_{S_i}$ can be determined. The propagation of robustness through $\phi$ is reliant on $\min$ and $\max$ operators over relevant time intervals $I$, which can be encoded exactly in MILP optimization problems.
Stay-in behavior for a convex polytope $RoI$ can be encoded as
\begin{multline}
\label{eq:opt_free_embedding}
    x_{S} \in RoI \impliedby \big( \bigwedge_{f=1}^{n^{\textrm{faces}}} \bigwedge_{j=0}^{d} (H^{f} r(i)^{(j)} \leq b^{f}), \\
    \forall i \in \mathbb{N}_{[0,N-1]} \big),
\end{multline}
where $n^{\textrm{faces}}$ denotes the number of faces of the $RoI$ polytope in the linear inequality form, $Hx \leq b$ and $d$ denotes the degree of the B\'ezier curve.
Additionally, collision avoidance with $RoI$ can be encoded by ensuring that all control points of a B\'ezier curve are outside at least one surface of $RoI$
\begin{multline}
\label{eq:opt_col_embedding}
    x_{S} \notin RoI \impliedby
    \big( \bigvee_{f=1}^{n^{\textrm{faces}}} \bigwedge_{j=0}^d (H^f r(i)^{(j)} \geq b^f ), \\
    \forall i \in \mathbb{N}_{[0,N-1]} \big),
\end{multline}
where the disjunctions over the faces of the objects are encoded via the well-known big-M method.

\subsubsection{Dynamics and Impacts}
The integrator constraint in Eq.~\eqref{eq:opt_control_R} and~\eqref{eq:opt_control_O} is apparent via the construction of the trajectories in Eq.~\eqref{eq:bezier_formulation}. For modeling of the impact kinematics, we keep track of variables $z_{(R,O)} \in \mathbb{B}^{(N_{R}\times N_{O})}$ indicating whether curve $N_i$ of robot $R$ has an impact with $N_j$ of object $O$. We can use these variables to embed conditional smoothness and impact conditions
\begin{multline}
\label{eq:opt_dyn_embedding}
    \dot{x}_R^{-} = \dot{x}_R^{+} \impliedby 
    \big(\dot{h}^{(0)}_{R}(i+1) = \dot{h}^{(d-1)}(i) \wedge \\
    \dot{r}^{(0)}(i+1) = \dot{r}^{(d-1)}(i), \forall i \in \mathbb{N}_{[0,N-2]},\\
     \iff z_{(R,O)}[i,j] = \bot, \forall O \in \mathcal O, \forall j \in N_O),
\end{multline}
indicating that this smoothness constraint is only enforced if this robot's curve does not intersect with any curves of any object. 
In a similar conditional manner, we embed the impact conditions
\begin{multline}
\label{eq:opt_dyn_embedding}
    \dot{p}_R^{+}(i) = m_1\dot{p}_R^{-} + m_2 \dot{p}_O^{-} \impliedby \\
    \big(\dot{h}^{(0)}_{R}(i+1) = m_1\dot{h}^{(d-1)}_{R}(i) + m_2\dot{h}^{(d-1)}_{O}(j) \wedge \\
    \big(\dot{r}^{(0)}_{R}(i+1) = m_1\dot{r}^{(d-1)}_{R}(i) + m_2\dot{r}^{(d-1)}_{O}(j), \\
    \forall i \in \mathbb{N}_{[0,N-2]}, \iff z_{(R,O)}[i,j] = 1 \big),
\end{multline}
where, for the Impact-Robust planner, this is performed $\forall v \in V(\hat{Z}_{O,X_0})$ and the robustness term $\delta$ is added.

\subsubsection{Velocity Bounds}
Velocity constraints in Eq.~\eqref{eq:opt_vel} can be encoded due to the bounding-box property of the control points of B\'ezier curve $i$ of system $S$
\begin{multline}
\label{eq:opt_vel_embedding}
    \dot{x}_{S} \in \mathcal{V} \impliedby \big( \dot{h}(i)^{(j)}\underline{\dot{x}}_{S} \leq \dot{r}(i)^{(j)} \leq \dot{h}(i)^{(j)}\bar{\dot{x}}_{S}, \\
    \forall i \in \mathbb{N}_{[0,N-1]}, \forall j \in \mathbb{N}_{[0,\ldots,d-1]} \big),
\end{multline}
where $\underline{\dot{x}}_{S}$ and $\bar{\dot{x}}_{S}$ are the velocity lower- and upper-bound given by the user or physical properties of the robot.

\subsubsection{Collision Avoidance}
    As trajectories are parametrized by B\'ezier curves or Zonotopes, collision checks between a robot at time $t$ and an object at time $t$ relies on checking the intersection of the entire curve for which $t \in [h^{(0)}_{R},h^{(d)}_{R}]$ and $t \in [h^{(0)}_{O},h^{(d)}_{O}]$. This is in sharp contrast with the requirement that robots and objects should impact in order to generate motion of the objects\footnote{Specifically, unless either the robot or object velocity is zero and the restitution coefficient, $c$, is 1, post-impact bounding boxes are guaranteed to intersect, regardless if the physical systems do.}. A heuristic \emph{best-effort} collision avoidance approach is to indicate pre- and post-impact B\'eziers and not constrain collisions on these curves.

We found however that this led to significant additional complexity in the MILP as well as requiring additional heuristics. To this end, we only ensure collision avoidance between robots and between objects and defer any collision avoidance between robots and objects to safety constraints in the controller.
If the trajectories of $S_i$ and $S_j$ are parametrized by B\'ezier curves (e.g. objects and robots in the best-case planner or robots in the impact-robust planner), we can ensure collision avoidance with
\begin{multline}
    ||p_{S_i}(t)-p_{S_j}(t)||_2 \leq \textrm{rad}_{S_i} + \textrm{rad}_{S_j}, \forall t \impliedby \\
    \big( \textrm{box}(r_{S_i}(k)) \oplus \mathcal{B}_{\infty}^{\textrm{rad}_{S_i}} \cap \textrm{box}(r_{S_j}(l)) \oplus \mathcal{B}_{\infty}^{\textrm{rad}_{S_j}} = \varnothing, \\
    \forall k,l \in \mathbb{N}_{[0,N-1]}, \iff h(k) \cap h(l) \neq \varnothing \big),
\end{multline}
where $\mathcal{B}_{\infty}^{\textrm{rad}}$ denotes the infinity-norm ball with radius $\textrm{rad}$. If $S_i$ and $S_j$ are parametrized by zonotopes (e.g. objects in the impact-robust planner), we can ensure collision avoidance with
\begin{multline}
    \label{eq:opt_col_objects_embedding}
        ||p_{S_i}(t) - p_{S_j}(t)||_2 \geq r_{S_i} + r_{S_j}, \forall t \impliedby \\
        \big( \textrm{box}(\hat{Z}^{+}_{S_i}(k),\hat{Z}_{S_i}^{-}(k)) \oplus \mathcal{B}_{\infty}^{\textrm{rad}_{S_i}} \cap \textrm{box}(\hat{Z}^{-}_{S_j}(l),\hat{Z}_{S_j}^{+}(l)) \oplus \mathcal{B}_{\infty}^{\textrm{rad}_{S_j}} \\
        \forall k,l \in \mathbb{N}_{[0,N-1]}, \iff h(k) \cap h(l) \neq \varnothing \big),
\end{multline}
with the zonotope interval hull overapproximation $\hat{Z}_{S_i} \supseteq Z_{v,S_i}$ via Eq.~\eqref{eq:interval_hull}.
All collision avoidance equations are therefore sound but conservative.

\subsubsection{Acceleration Attenuation}
As mentioned in Sec.\ref{sec:problem_formulation}, while the B\'ezier trajectories can be arbitrarily smooth and adhere to position and velocity constraints, any constraint on higher-order derivatives leads to non-convex equations (according to the derivative of Eq.~\eqref{eq:bezier_formulation}). As such, we add the potential to attenuate acceleration by the quadratic cost
\begin{equation}
\label{eq:cost_attenuation}
    J(r_{S_i},h_{S_i}) = \sum_{k=0}^{N-1} \ddot{r}_{S_i}(k)^T Q_{\ddot{r}} \ddot{r}_{S_i}(k) + \ddot{h}_{S_i}(k)^T Q_{\ddot{h}} \ddot{h}_{S_i}(k),
\end{equation}
as described in \cite{marcucci2023motion}. The introduction of this cost term makes what was a Mixed-Integer Linear Program into a Mixed-Integer Quadratic Program, adding computational complexity.

\section{Online Two-Body Impact Planner}
\label{sec:replanner}
A significant assumption in the offline Mixed-Integer Problem (MIP) planner in Sec.~\ref{sec:implementation} is that collisions happen between point masses. 
The reason for this assumption is that for point masses, all impact conditions are linear, enabling globally optimal solutions with spatio-temporal tasks via mixed-integer programming. 
However, upon execution of the plan, the physical sizes of the robots and objects need to be taken into account via Eq.~\eqref{eq:robot_cylinder_impact}. 
Additionally, changes in the object position and velocity require careful replanning in order to generate the desirable post-impact velocities obtained from the MILP planner. 

To that end, we consider an online planner that, given the currently observed position and velocity of an object, and the desired post-impact velocity of the object, recomputes the B\'ezier curves before and after impact. We enable the replanner when the robot is on a pre-impact B\'ezier curve and the object is on a free floating segment towards that impact.
We formulate the replanning via a Quadratic Program (QP)
\begin{subequations}
\label{eq:opt2}
\begin{align}
    \min_{r_{R},h_{R}}& \hquad \ddot{r}_{R}(s)^T Q_{\ddot{r}} \ddot{r}_{R}(s) + \ddot{h}_{R}(s)^T Q_{\ddot{h}}\ddot{h}_{R}(s) + \delta_{p} + \delta_{\dot{p}}, \tag{\ref*{eq:opt2}}\\
    \textrm{s.t.} \hquad & [x_{R}(t_0),\dot{x}_{R}(t_0)] = [x^{\textrm{MILP}}(t_0),\dot{x}^{\textrm{MILP}}(t_0)], \label{eq:opt2nit} \\
    & [x_{R}(t_f),\dot{x}_{R}(t_f)] = [x^{\textrm{MILP}}(t_f),\dot{x}^{\textrm{MILP}}(t_f)], \label{eq:opt2_final}\\
    & ||x_{R}(t_{\textrm{impact}}) - {x}_{R}^{\textrm{des}}(t_{\textrm{impact}})||_2^2 =  \delta_{p}, \label{eq:opt2_radius}\\
    & ||\dot{x}_{R}(t_{\textrm{impact}}) - \dot{x}_{R}^{\textrm{des}}(t_{\textrm{impact}})||_2^2 = \delta_{\dot{p}}, \label{eq:opt2_coll} \\
    & \textrm{Eqs. } \eqref{eq:opt_vel}, \label{eq:opt2_others}
\end{align}
\end{subequations}
where $t_0$, $t_f$, $t_{\textrm{impact}}$ are the start of the pre-impact B\'ezier, the end of the post-impact B\'ezier and the planned impact time respectively, precomputed by the offline planner.
We obtain $x_{R}^{\textrm{des}}$ and $\dot{x}_{R}^{\textrm{des}}$ from solving Eq.~\eqref{eq:cylinder_impact_rotated} (see Appendix.~\ref{ssec:two-body-impact-problem}) using predicted pre-impact states of the object
\begin{align*}
    \tilde{x}_{O}(t_{\textrm{impact},i}) &= \tilde{x}_{O}(t) + \dot{\tilde{x}}_{O}(t)(t_{\textrm{impact,i}}-t),\\
    \dot{x}_{O,i}^{\textrm{des}} &= \frac{x_{O,i+1}^{\textrm{des}} - \tilde{x}_{O}(t_{\textrm{impact},i})}{t_{\textrm{impact},i+1} - t_{\textrm{impact},i}},
\end{align*}
where $\tilde{x}_{O}(t)$ and $\dot{\tilde{x}}_{O}(t)$ are estimated positions and velocities of the object at the current time of replanning $t$, $\tilde{x}_{O}(t_{\textrm{impact},i}$ denotes the estimated position of the object at the impact time, $x_{O,i+1}^{\textrm{des}}$ denotes the desired position of the object at the next impact from the offline planner, and $\dot{x}_{O,i}^{+}$ is the desired position of object $O$ at impact $i$ to ensure this behavior. It is $\dot{x}_{O,i}^{+}$ which is one the left-hand side of Eq.~\eqref{eq:cylinder_impact_rotated}.
Cost terms~\eqref{eq:opt2_radius} and~\eqref{eq:opt2_coll} then ensure that at time $t_{\textrm{impact}}$, the robot impacts to ensure minimal deviation from the solution of Eq.~\eqref{eq:cylinder_impact_rotated}.

Although the problem in Eq.~\eqref{eq:cylinder_impact_rotated} is non-convex, the offline MIP planner provides excellent initial guesses on $\theta$ and $\dot{p}_{S_i}^{+}$.
In contrast to the offline MIP planner, this recomputed online plan considers the physical sizes of objects and robots as shown in Fig.~\ref{fig:impact_example}.
For validation, we include workspace constraints in Eq.~\eqref{eq:opt2} via Eq.~\eqref{eq:opt_free_embedding}.

\begin{remark}
    Not considering $\phi$ in Eq.~\eqref{eq:opt2} means that STL satisfaction is not guaranteed upon execution. 
    Based on the reference trajectories from the offline planner, the execution realizes desired pre-impact positions and velocities of the objects at the next impact, at the cost of accurate tracking of the B\'ezier curves of the robots, whether or not satisfaction of $\phi$ is dependent on bounded-error tracking of the original planned trajectories.
\end{remark}

\begin{figure*}[t]
    \centering
    \includegraphics[width=0.98\textwidth]{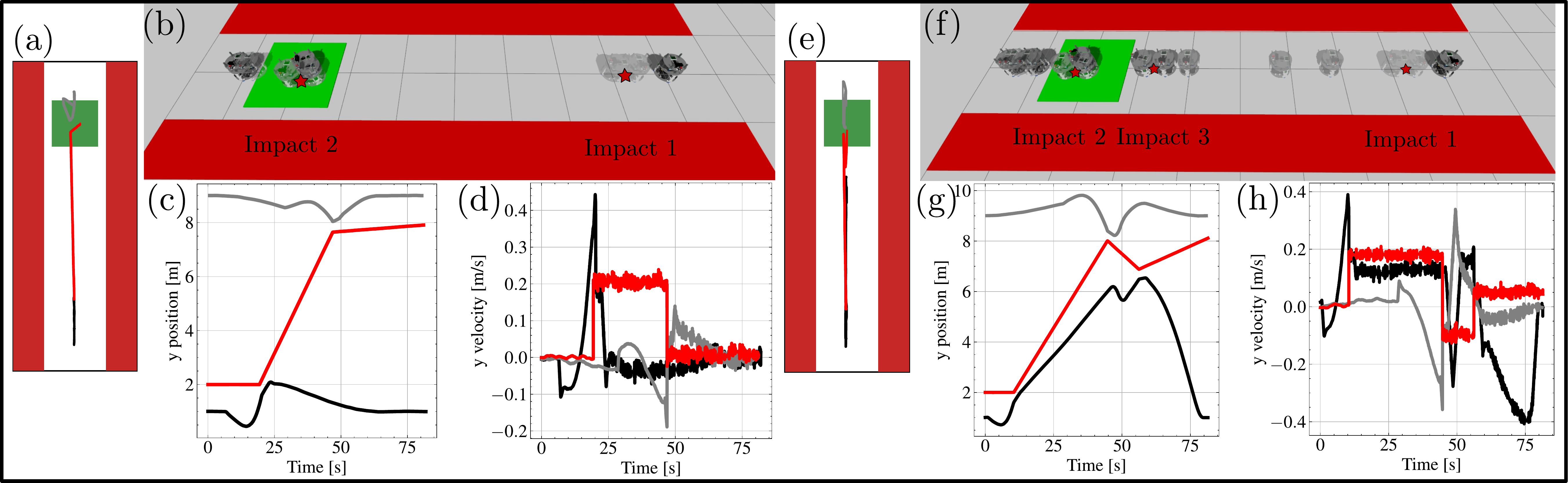}
    \caption{Simulation results of the corridor travel scenario. 
    The initial setup and planned trajectory are shown in Fig.~\ref{fig:corridor_space_robust} and~\ref{fig:corridor_impact_robust} respectively.
    On the left is the spatially robust- and on the right the impact-robust planner result.
    (a) the executed trajectory. (b) snapshots of the simulation where the robots have moved from low to high opacity instances. The impacts are highlighted by the red stars. (c) the vertical position over time with the red dotted line indicating the preplanned object position. (d) the vertical velocity over time. Note that the velocity change of $O_1$ is instantaneous due to the impact with $R_1$ and $R_2$.}
    \label{fig:worst_case_best_case_experiments}
\end{figure*}

\section{Impact-Aware MPC}
\label{sec:implementation_impact_aware_mpc}
We now consider the real-time control of the robots such that Eq.~\eqref{eq:opt_prob} may be solved on physical free-flying platforms.
First, we consider the following rigid body force and rate model of a free-flyer
\begin{subequations}
\label{eq:body_force_torque_mpc_model}
\begin{align}
    \dot{p} &= v, \\
    \dot{v} &= \frac{1}{m}R(q)^T F, \\
    \dot{q} &= \frac{1}{2}\Xi(q)\tau,
\end{align}
\end{subequations}
where the control input vector $u$, for the planar case, is defined as $u=[F^T,\tau^T]^T \in \mathbb{U}\subset\mathbb{R}^3$ where $F \subset \mathbb{R}^2$ and $\tau \subset \mathbb{R}^1$ are applied forces and torques in the body frame. 
We refer to App.~\ref{sec:appendix} for more details on the robot properties.
For the object, we consider Eq.~\eqref{eq:body_force_torque_mpc_model} without actuation, e.g., $\dot{p}=v$, $\dot{q}=\frac{1}{2}\Xi(q)\omega$ with equivalent model properties.

The low-level MPC that controls the robots ensures control actuation minimization, accurate tracking, collision avoidance constraints, and the desired post-impact velocities of objects.
We define the low-level Nonlinear MPC problem as
\begin{subequations}
\label{eq:MPC_formulation}
\begin{align}
J^{*}( \tilde{x}(& k)) = \min_{\mathbf{u}_k} \ J(x(n|k), u (n|k)), \tag{\ref*{eq:MPC_formulation}} \\
 \text{s.t.} \hspace{2mm} & x(m+1|k) = g(x(m|k), u(m|k)), \label{eq:mpc_set_cond0}\\
 & x(m|k)  \in \mathbb{X}, \label{eq:mpc_set_cond1}\\
 & u(m|k) \in \mathbb{U}, \ \forall m\in\mathbb{N}_{[0,N-1]}, \label{eq:mpc_set_cond3} \\
 & x(0|k) = x(k). \forall n\in\mathbb{N}_{[0,N]},
\end{align}
\end{subequations}
where the discrete-time dynamics, $g(x(m|k),u(m|k))$, consider the impact according to
\begin{equation*}
    g(x(m|k),u(m|k)) = \begin{cases}
        \begin{bmatrix} \textrm{Eq.~\ref{eq:cylinder_impact_rotated}} \end{bmatrix}^T, &\textrm{if } t_m = t_{\textrm{impact}} \\
        \textrm{RK4} (\begin{bmatrix}\textrm{Eq.~\ref{eq:body_force_torque_mpc_model}} \end{bmatrix}^T), &\textrm{otherwise}
    \end{cases}
\end{equation*}

where $\textrm{RK4}$ denotes the integration of the continuous dynamics in Eq.~\eqref{eq:body_force_torque_mpc_model} with the Runge-Kutta 4 scheme.
The cost function, $J(x(n|k),u(n|k))$ for B\'ezier tracking is detailed in Appendix.~\ref{ssec:appendix_mpc_details}.
The MPC scheme ensures accurate tracking of the (re)planned B\'ezier curves under the planned impact occurrence. 
An impact detector is used to ensure that the pre- or post-impact stage of a robot w.r.t. the object is accurately considered in the controller and online planner.
We change the weights in $Q$ and $Q_N$ based on whether the impact is in the MPC horizon and whether the state $k$ is before or after the impact. Details are shown in Appendix.~\ref{ssec:appendix_mpc_details}.

%% file: 6_Results.tex
\begin{figure*}[t]
    \centering
    \includegraphics[width=0.98\textwidth]{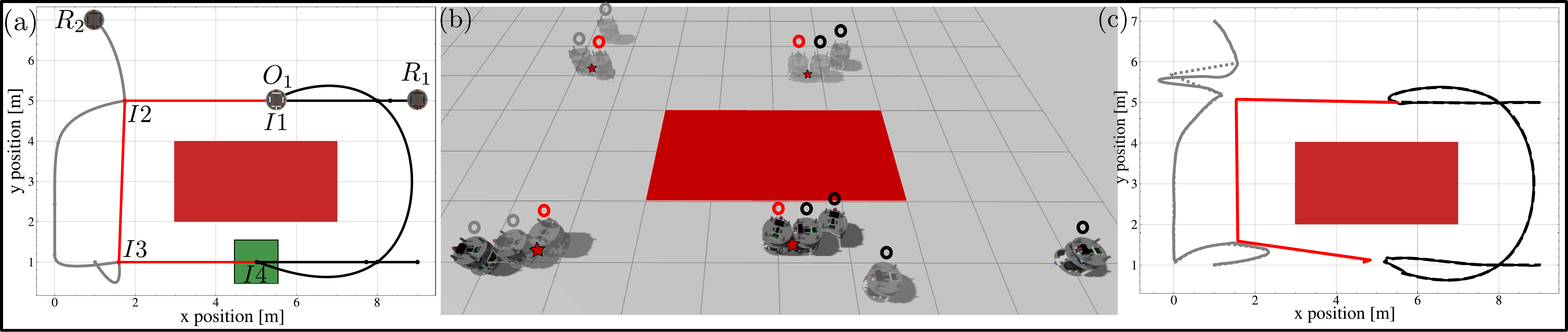}
    \caption{Obstacle avoidance scenario where object $O_1$ needs to be moved from the top of the red obstacle to its bottom. (a) the preplanned trajectory with the impact-robust MILP planner, $\rho_{\phi} = 0.5m$. Four impacts are planned with the initial and final impact with robot $R_1$ on the top right and bottom right respectively. (b) snapshots from the simulator. The impacts are highlighted by the red stars. (c) the executed trajectories, indicating the replanned trajectories to accommodate the physical sizes of the robots and object and their updated state.}
    \label{fig:results_sim_obstacle_avoidance_bestcase}
\end{figure*}

\begin{figure*}[t]
    \centering
    \includegraphics[width=0.98\textwidth]{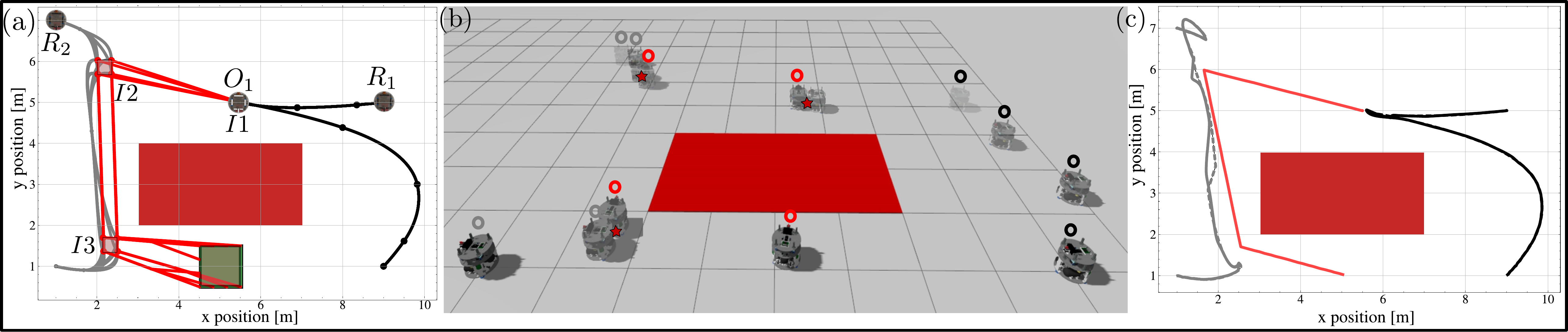}
    \caption{Obstacle avoidance scenario where object $O_1$ needs to be moved from the top of the red obstacle to its bottom. (a) the preplanned trajectory with the impact-robust MILP planner, $\delta = 0.0055 \frac{m}{s}$. Three impacts are planned. The post-impact uncertainty over the first impact makes the deterministic initial state an uncertain state for the next impact, denoted by a zonotope interval hull $\hat{Z}_{O_1}$ (b) snapshots from the simulator. The impacts are highlighted by the red stars. (c) the executed trajectories, indicating the replanned trajectories to accommodate the physical sizes of the robots and object and their updated state.}
    \label{fig:results_sim_obstacle_avoidance_worstcase}
\end{figure*}


\section{Simulations and experiments}
\label{sec:results}

In this section, we present high-fidelity simulator results and real-world experiments on the ATMOS freeflyer platform~\cite{roque2025towards}. 
Due to the fact that in our simulator experiments we run the entire software stack existing on the robot (Software-In-The-Loop, or SITL), we use the MPC weights in Sec.\ref{ssec:appendix_mpc_details} for both simulations and experiments.
Position and attitude measurements from either Gazebo or a motion capture system are fused with IMU data using an EKF onboard the low-level PX4 flight controller.
After solving the MIP with Gurobi~\cite{gurobi}, all replans and MPCs are solved using Casadi~\cite{andersson2019casadi} run centrally on a laptop with an Intel Core i7-14700HX CPU and 32Gb of RAM.
The MPC sends force and rate commands to the low-level PX4 flight controller, either physically on the robot over a Wifi6 network or on a simulated PX4 on the laptop.
For all scenarios we introduce the quadratic acceleration attenuation cost term in Eq.~\eqref{eq:cost_attenuation}.

Code and Videos at \url{https://joris997.github.io/impact_stl/}.

\subsection{Simulator Results}
\label{ssec:results_simulator}
We perform simulations in the high-fidelity Gazebo simulator. 
We simulate the entire software stack of the freeflyer robots used later in the experimental results. 
For two scenarios, we qualitatively compare the model-free spatially robust plan and the impact-robust plan. 
From the impact-robust plan, which has B\'ezier curves for all outer vertices of the zonotope representing the state of the object, we interpolate these B\'ezier curves to obtain a nominal motion plan.

\subsubsection{Corridor Scenario}
First, we consider the running corridor example where object $O_1$ is to be transported between two robots.
The planned trajectories for this transportation scenario have already been presented in Fig.~\ref{fig:corridor_space_robust} and~\ref{fig:corridor_impact_robust} for the spatially robust and impact-robust optimization metrics respectively.
We define $R_1$, $R_2$ and $O_1$ as the two robots and object with initial and final conditions: $R_1(0)=[1,1], R_1(100)=[1,1]$, $R_2(0)=[1,9], R_2(100)=[1,9]$ and $O_1(0)=[1,2]$. 
The overall specification is for the object to reach a goal region at 100 sec. and for all systems to avoid obstacles.
\begin{equation*}
    \phi = \Box_{[100,100]}(p_{O_1} \in \textrm{goal}) \bigwedge_{S_i \in \mathcal{S}} \Box_{[0,100]}(p_{S_i} \in \mathcal{W}_{free}),
\end{equation*}
where the goal region is shown in Fig.~\ref{fig:corridor_space_robust} and~\ref{fig:corridor_impact_robust}. 

For the \textbf{Spatially Robust Planner} the simulation results of Fig.~\ref{fig:corridor_space_robust} are shown on the left in Fig.~\ref{fig:worst_case_best_case_experiments}. Note the replanned trajectory, especially apparent for $R_1$ at $t\approx 5$. The point-mass assumption from the offline planner has to be relaxed and generating the desired pre-impact velocity of $R_1$ requires moving backwards to get more distance between $R_1$ and $O_1$. Notice that $R_2$ also replans and, as shown on the most-left sub-figure, counteracts the horizontal deviation of $O_1$ by a horizontal deviation in the impact.

For the \textbf{Impact Robust Planner} the simulation results of Fig.~\ref{fig:corridor_impact_robust} are shown on the right in Fig.~\ref{fig:worst_case_best_case_experiments}. Notice that after the second impact, $O_1$ reaches the goal with a non-zero velocity. In contrast to the spatially robust planner, where the non-zero velocity was not planned, here this post-impact velocity is specifically taken into account w.r.t. the satisfaction of $\phi$ (indicated by the solid red line being between the dashed red lines). 
The maximal uncertainty in the post-impact velocity is $\delta=0.019 \frac{m}{s}$ denoting that after any impact that occurs, a positive or negative deviation of $0.019$ meters per second may occur (both in the $x$ and $y$ direction).

\subsubsection{Obstacle Avoidance Scenario}
We now consider the specification to bring an object from an initial state to a final state via impacts with two other robots. The environment is shown in Fig.~\ref{fig:results_sim_obstacle_avoidance_bestcase}. Consider again $R_1$, $R_2$, and $O_1$  with initial and final condition: $R_1(0)=[9,5]$, $R_1(150)=[9,1]$, $R_2(0)=[1,7]$, $R_2(150)=[1,1]$, and $O_1(0)=[5.5,5]$. 
The overall specification is
\begin{equation*}
    \phi = \diamondsuit_{[0,150]}(p_{O_1} \in \textrm{goal}) \bigwedge_{S_i \in \mathcal{S}} \Box_{[0,150]}(p_{S_i} \notin Obs),
\end{equation*}
where the goal and obstacle are shown in Fig.~\ref{fig:results_sim_obstacle_avoidance_bestcase} and~\ref{fig:results_sim_obstacle_avoidance_worstcase}.

For the \textbf{Spatially Robust Planner} in Fig.~\ref{fig:results_sim_obstacle_avoidance_bestcase}, notice that both the robot and the objects try to maximize the distance to the wall and that the object proceeds towards the center of the goal region. Notice however, that the distance to the wall is larger than at the initial position of the object as the $\Box_{[0,150]}$ considers the minimal distance in the time-horizon.
Upon execution, a significant replan w.r.t. the offline plan is necessary to steer the object from top left to bottom left. This is due to the fact that for point masses with $e=1$, the velocities will simply swap (according to Fig.~\ref{fig:parametrization}). However, for the physical bodies and the real restitution coefficient on the robot, a significant stopping force is required to remove the horizontal velocity of~$O_1$.

\begin{figure*}[t]
    \centering
    \includegraphics[width=0.98\textwidth]{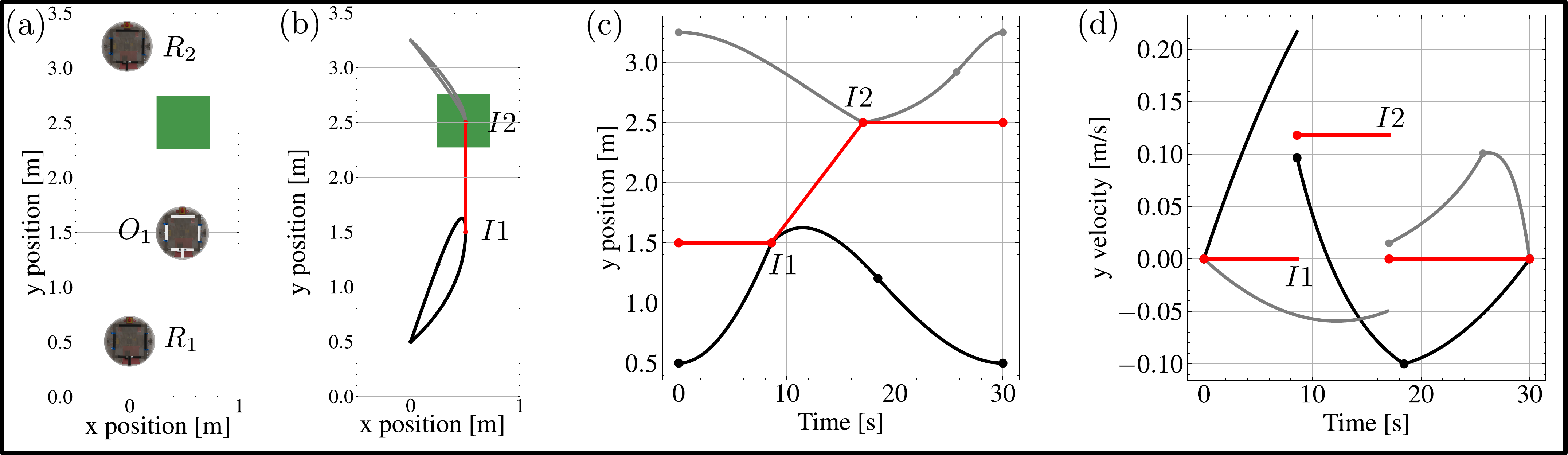}
    \caption{Simple object transportation scenario where $O_1$ should be moved to the green region, starting from the initial setup in (a). (b) the preplanned trajectories with the spatially robust planner, $\rho_{\phi} = 0.25m$. (c) vertical position over time. (d) the vertical velocity over time.}
    \label{fig:lab_test_2}
    \vspace{-0.4cm}
\end{figure*}

\begin{figure}[t]
    \centering
    \includegraphics[width=0.48\textwidth]{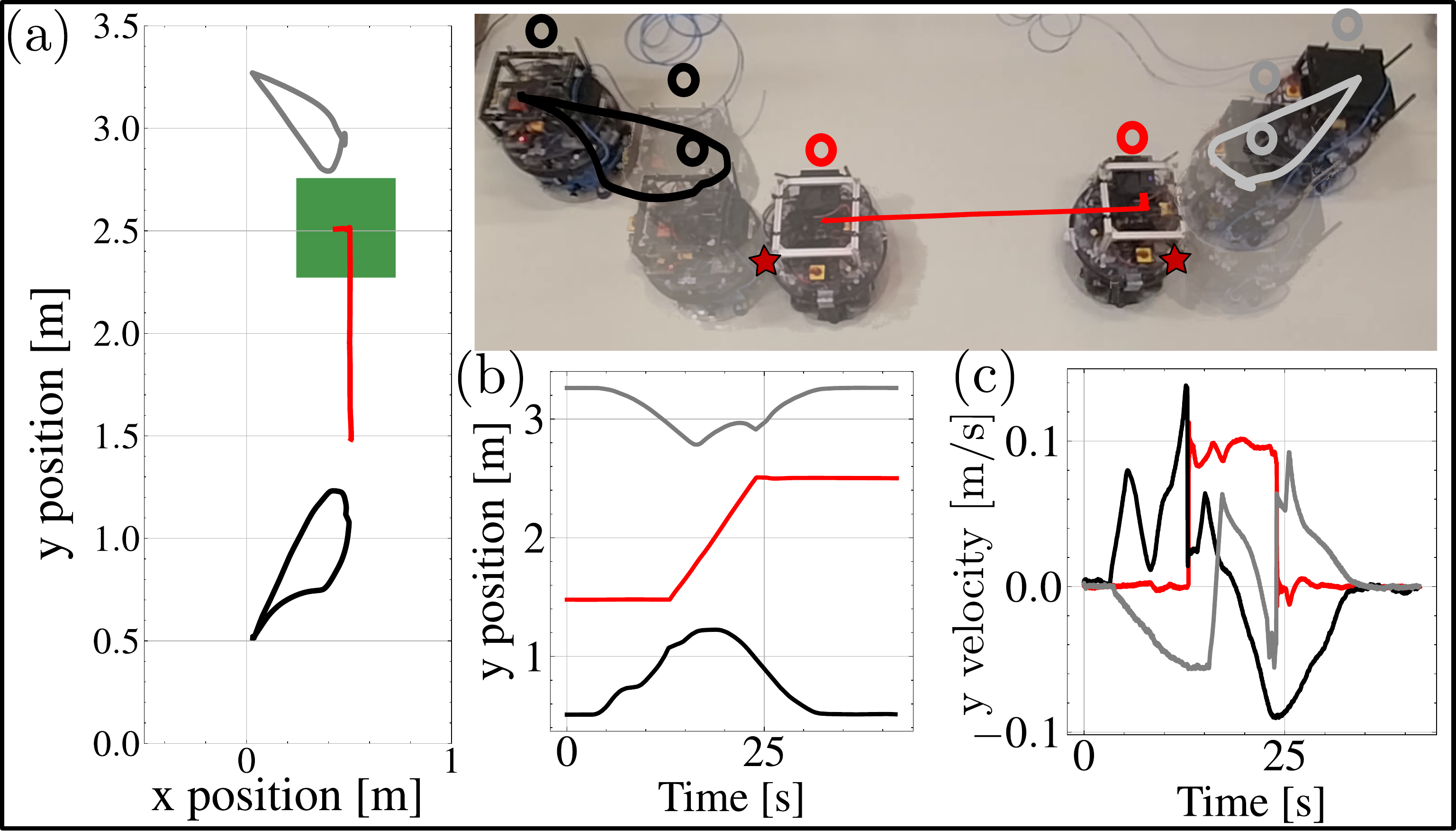}
    \caption{Experimental results of the simple object transportation scenario in Fig.~\ref{fig:lab_test_2}. (a) the executed replanned trajectories from the online replanner in Sec.~\ref{sec:replanner}. (b) the vertical position over time, notice the deviation from the nominal plan in Fig.~\ref{fig:lab_test_2}c. (c) the vertical velocity over time.}
    \label{fig:lab_test_2_exp}
    \vspace{-0.5cm}
\end{figure}

\begin{figure*}[t]
    \centering
    \includegraphics[width=0.98\textwidth]{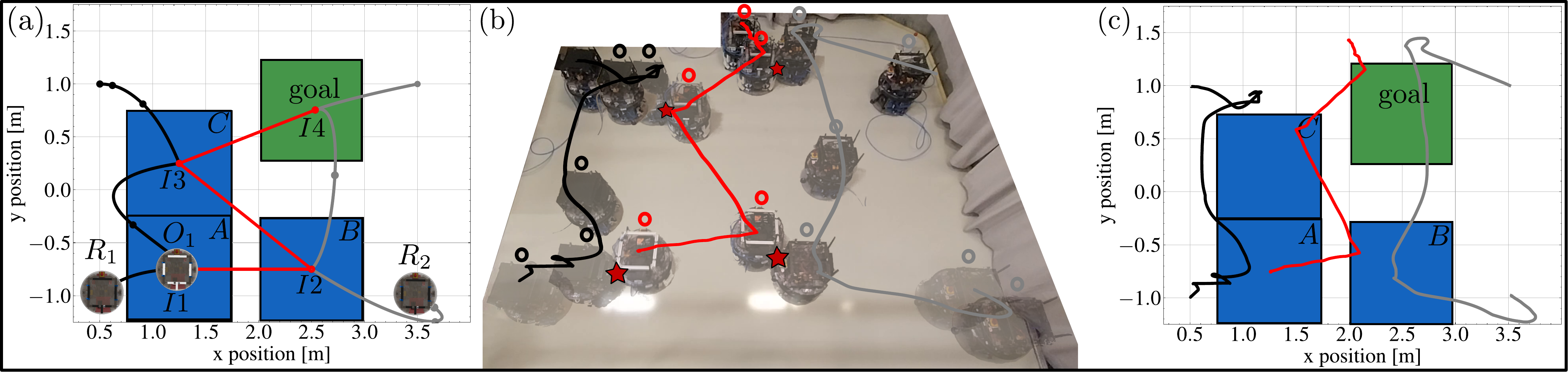}
    \caption{Complex Pong scenario where $O_1$ should visit $A$, $B$, and $C$, but it is not pre-specified in which order this should be. (a) the preplanned trajectories from the spatially robust planner (with an additional cost term for acceleration attenuation) (b) snapshots from the hardware experiment. The impacts 4 planned impacts are denoted by red stars. (c) the realized positions w.r.t. the specification $\phi$. The spatial deviations make the specification difficult to satisfy, motivating the robustness metrics.}
    \label{fig:pingpong_exp}
\end{figure*}

For the \textbf{Impact Robust Planner} in Fig.~\ref{fig:results_sim_obstacle_avoidance_worstcase}, notice that the offline planner plans three impacts. It is apparent that the additional fourth impact in the goal region for the spatially robust planner will not affect the overall robustness $\delta = 0.0055 \frac{m}{s}$ as the limiting factor is the vertical traversal. Note in the executed trajectory that $R_2$ needs to overtake $O_1$, with which it has impacted. A Control Barrier Function ensures safety during these kind of maneuvers.



\subsection{Experimental Results}
\label{ssec:results_experiments}
We now perform hardware validation on the freeflyer platform. 
The platform floats via three passive air-bearings located beneath the robot and is (ignoring air friction) weightless in the plane. 
It is actuated by 8 solenoid air valves, actuated at a frequency of 10Hz. For details on the entire experimental setup, we refer to~\cite{roque2025towards}.

Although the platforms are not subject to any significant friction due to its floating and low velocities, any unevenness in the floor will significantly alter a free-floating object's trajectory. To that end, we run a velocity-keeping MPC on the object, detailed in Appendix.~\ref{ssec:appendix_mpc_velocity}.

\subsubsection{Throw-and-Catch Scenario}
We perform a simple throw-and-catch scenario, not unlike the corridor scenario we performed in simulation. For clarity, we consider an initial state of the robots that is different from the object in both directions.
Consider again $R_1$, $R_2$, and $O_1$ with initial and final condition: $R_1(0)=[0,0.5]$, $R_1(30)=[0,0.5]$, $R_2(0)=[0,3.25]$, $R_2(30)=[0,3.25]$, and $O_1(0)=[0.5,1.5]$. 
The overall specification is
\begin{equation*}
    \phi = \Box_{[30,30]}(p_{O_1} \in \textrm{goal}),
\end{equation*}

The \textbf{Spatially Robust Planner} results are shown in Fig.~\ref{fig:lab_test_2} and the experimental results in Fig.~\ref{fig:lab_test_2_exp}. Notice that, similar to the corridor scenario with the spatially robust planner, two impacts are planned that \emph{throw} and \emph{catch} the object. 
Again, replanned trajectories take a different approach that the nominal plan to account for the physical sizes of the robot.
Notice in the velocity plot in Fig.~\ref{fig:lab_test_2_exp} that the post-impact velocity of $O_1$ is less noisy but deviates slightly. This is due to the velocity-keeping MPC on $O_1$. To ensure an accurate replan for the \emph{catch} by $R_2$, we estimate $O_1$'s velocity over a pre-specified duration (4 seconds in this experiment).

\subsubsection{Pong Scenario}
We consider a more complex scenario where an object should visit multiple areas of interest consecutively. The initial and final condition are: $R_1(0)=[0.5,-1.0]$, $R_1(100)=[0.5,1.0]$, $R_2(0)=[3.5,-1]$, $R_2(100)=[3.5,1]$ and $O_1(0) = [1.25,-0.75]$. 
The overall specification is
\begin{align*}
    \phi = & \diamondsuit_{[0,100]}(p_{O_1} \in B) \land \diamondsuit_{[0,100]}(p_{O_1} \in C) \land \\ & \Box_{[100,100]}(p_{O_1} \in D),
\end{align*}

The \textbf{Spatially Robust Planner} results are shown in Fig.~\ref{fig:pingpong_exp}. Note that the offline motion planner resolves the ambiguity in which order to satisfy $\diamondsuit_{[0,100]}(p_{O_1} \in C)$ and $\diamondsuit_{[0,100]}(p_{O_1} \in D)$ based on the location of the areas and the robots. As we maximize the spatial robustness of $\phi$, $O_1$ plans to visit each region in its center. 
The execution shows the significant alterations to the preplanned motion plan, highlighting the necessity of the replanner.

We notice a significant deviation in the planned trajectory for $O_1$. 
We hypothesize that this is caused by using a constant restitution coefficient and that preplanned impact times are held constant. 
Extending the impact time after the first impact could ensure that $O_1$ achieves a greater spatial robustness. 

%% file: 7_Limitations.tex
\section{Limitations}
\label{sec:limitations}
Our impact-aware planner, replanner, and MPC is planning robust trajectories while considering online adaptation to changing circumstances. 
However, it is not without its limitations, and we highlight several here.

First, while the online replanner could take on different forms to consider different robot and object shapes, the current implementation is limited to 2D circular robots and objects.

Additionally, while the MILP planner is robust to uncertainty in the post-impact velocity of objects, this is only with respect to the idealized kinematic impact model.
The model-based robustness metric $\delta$ therefore only relates to an abstraction of the true model-based robustness. In reality, there are complex impact dynamics at play that we do not consider in the MILP planner or in the replanner. 
The effect of this is apparent in the Pong scenario in Fig.~\ref{fig:pingpong_exp}.

Thirdly, the replanner does not consider the specification $\phi$ in its replanning strategy. 
Doing this could ensure that the updated plan is minimally violating \cite{sadraddini2015robust} but it would reintroduce binary variables (or non-convexities \cite{gilpin2020smooth}) leading to a centralized formulation that is too computationally expensive to reliably update on the fly.
Now, robots may unnecessarily violate their specification, $\phi_{R_i}$, to accommodate a post-impact velocity for an object, determined by the offline planner.

Lastly, the B\'ezier parametrization for the robots is unable to provide dynamically feasible trajectories as acceleration constraints would be non-convex \cite{von2024using}. 
Instead we indirectly penalize accelerations via the cost term in Eq.~\eqref{eq:opt2}. 
We further limit dynamic feasibility concerns via lower-bounding $\dot{h}_{R_i}$ and conservative velocity constraints $\mathcal{V}_{\mathcal R}$.


%% file: 8_Conclusion.tex
\section{Conclusion}
\label{sec:conclusion}

In this paper, we have proposed a solution to the complex problem of collaborative transportation of passive objects in space.
We have shown how the lack of gravity can be effectively utilized for synthesizing minimal interaction transportation by instantaneously changing an object's velocity via impacts with controllable robots.
The absence of gravity then allows for passive floating through space, until another impact changes its velocity.
We have also shown how an impact-robustness metric is able to implicitly generate desirable behavior when uncertainty in impact dynamics are considered.
We have validated the conjoined offline planning, online planning, and low-level control approach in high-fidelity simulation examples and on hardware.

Future work will involve generalizations to objects and robots of different shapes, replanning with velocity matching instead of impacts (for fragile objects) and replanning while considering minimally violating the high-level specification. Additionally, we wish to address the scenario of on-orbit servicing by considering orbital dynamics in our approach.
An additional avenue is the use of online impact parameter estimation to iteratively refine the kinematic impact model.

\section{Acknowledgements}
\label{sec:acknowledgements}
This work was partially supported by the Wallenberg AI, Autonomous Systems and Software Program (WASP) funded by the Knut and Alice Wallenberg Foundation. The authors are also affiliated with Digital Futures.
The authors would like to thank Pedro Roque, Elias Krantz, and Tafarrel Firhannoza Pramono for their help with the experimental validation.

%% file: Appendix.tex
\section{Appendix}
\label{sec:appendix}

\subsection{Two-body Impact Problem}
\label{ssec:two-body-impact-problem}
During replanning, we wish to consider the physical sizes and updated positions of the robots and objects. To that end, we first solve the the two-body impact problem from Eq.~\eqref{eq:robot_cylinder_impact} to obtain the desired pre-impact position and velocity of the robot, posed the minimum-norm optimization problem
\begin{subequations}
\label{eq:tbopt}
\begin{align}
    \min_{\substack{\dot{p}_{R,G}^{-},\dot{p}_{R,G}^{+},\\\dot{p}_{R,L}^{-},\dot{p}_{R,L}^{+}\\\dot{p}_{O,G}^{-},\dot{p}_{O,G}^{+},\\\dot{p}_{O,L}^{-},\dot{p}_{O,L}^{+}}}& \hquad ||\dot{p}_{O,G}^{+} - \dot{p}_{O,G}^{\textrm{des}}||_{Q_1} + ||\dot{p}_{R,G}^{-} - \dot{p}_{R,MILP}^{-}||_{Q_2} \tag{\ref*{eq:tbopt}}\\
    \textrm{s.t.} \hquad &\eqref{eq:robot_cylinder_impact} \\
    &\dot{p}_{R,L}^{-} = R(\theta)\dot{p}_{R,G}^{-} \\
    &\dot{p}_{R,L}^{-} = R(\theta)\dot{p}_{R,G}^{-} \\
    &\dot{p}_{R,G}^{+} = R^{-1}(\theta)\dot{p}_{R,L}^{+} \\
    &\dot{p}_{R,G}^{+} = R^{-1}(\theta)\dot{p}_{R,L}^{+}
\end{align}
\end{subequations}

\subsection{Impact-Aware MPC Details}
\label{ssec:appendix_mpc_details}
The cost function $J(x(n|k),u(n|k))$ in the impact-aware MPC is given as 
\begin{subequations}
\label{eq:mpc_cost}
\begin{align}
    J(\bar{x},x,u) &= \sum_{n=0}^{N-1} l\big(\bar{x}(n|k),x(n|k),u(n|k) \big) \nonumber \\
    & \qquad + V\big(\bar{x}(n|k),x(n|k)\big), \tag{\ref*{eq:mpc_cost}}\\
    l\big( \bar{x},x, u\big) &= \|e(n|k)\|^{2}_{Q} + \| u(n|k) \|^{2}_R, \\
    V\big(\bar{x},x\big) &= \| e(N|k) \|^{2}_{Q_N} \\ 
    e(k+n|k) &= \begin{cases}
      \bar{x}(n|k) - x(n|k), x\leftarrow \{p, v, \omega \}\\
      1 - (\bar{x}(n|k)^T x(n|k))^2, x \leftarrow  \{ q \} \end{cases}
\end{align}
\end{subequations}
where $\bar{x} \in \mathbb{R}^n$, $x \in \mathbb{R}^n$, and $u \in \mathbb{R}^m$ are the reference trajectory by evaluating the robot's B\'eziers Eq.~\eqref{eq:bezier_evaluating}, the state and the control input of $R_i$. $Q$, $R$, and $Q_N$ are positive definite weight matrices.
For all scenarios, we run the MPC with a horizon of 1 second and 10 sampling intervals.

\subsubsection{Weight-Scheduling}
While an impact traverses through the horizon of the MPC, we adapt the weights $Q$, $Q_N$ and $R$.
As we are mainly interested in creating the desired post-impact vector of the object (at the cost of potentially having the robot deviate from its post-impact pre-planned trajectory), we schedule the weights as follows
\begin{equation}
    Q = \begin{cases}
        \textrm{diag}([1e\text{-}3, 1e\text{-}3, 1e\text{-}3, \\ \hspace{10mm} 8e3, 8e3, 8e3, 8e3]) & \textrm{if } t_{impact} \in [t_{0},t_{N}] \\
        \textrm{diag}([5, 5, 5, \\ \hspace{10mm} 8e\text{-}1, 8e\text{-}1, 8e\text{-}1, 8e3]) & \textrm{otherwise}
    \end{cases} \notag
\end{equation}
\begin{equation}
    R = \textrm{diag}([1e\text{-}3, 1e\text{-}3, 1e\text{-}3, 2, 2, 2]) \notag
\end{equation}
\begin{equation}
    Q_N = 10 Q \notag
\end{equation}
with the $x=[p_x,p_y,p_z,\dot{p}_x,\dot{p}_y,\dot{p}_z,q]$ and $u=[F_x,F_y,F_z,\tau_x,\tau_y,\tau_z]$. 

\subsection{Velocity-Keeping MPC Details}
\label{ssec:appendix_mpc_velocity}
As the floor on the experimental setup is not completely flat, any object is not able to truly float freely between impacts. 
For online planning, deviations in the straight-line travel can be alleviated by using a height-map of the floor and forward integrating the observed state of the object.
However, for simplicity and a to make a true experimental version of the high-fidelity simulator, we instead opt to run a velocity-keeping MPC on the object. 

This MPC keeps position, detects an impact, turns off the controller for a pre-specified duration to estimate the post-impact velocity vector ($1$ second), and then draws a linear segment from the current state to a future state using this vector. 
\begin{equation}
    p_{ref}(t) = p \frac{1-t}{\Delta t} + (p+\Delta t\dot{\bar{p}}) \frac{t}{\Delta t}
\end{equation}
where $\Delta t >> 0$ is a duration that far exceeds the time to the next interval, $p$ is the position after the $1$ second duration, and $\dot{\bar{p}}$ is the averaged velocity over the $1$ second duration.
Afterwards, the MPC in Eq.~\eqref{eq:MPC_formulation} tracks this linear segment.

%% file: main.bbl
\begin{thebibliography}{33}
\providecommand{\natexlab}[1]{#1}
\providecommand{\url}[1]{\texttt{#1}}
\expandafter\ifx\csname urlstyle\endcsname\relax
  \providecommand{\doi}[1]{doi: #1}\else
  \providecommand{\doi}{doi: \begingroup \urlstyle{rm}\Url}\fi

\bibitem[Alonso-Mora et~al.(2017)Alonso-Mora, Baker, and Rus]{alonso2017multi}
Javier Alonso-Mora, Stuart Baker, and Daniela Rus.
\newblock \href{https://journals.sagepub.com/doi/full/10.1177/0278364917719333}{Multi-robot formation control and object transport in dynamic environments via constrained optimization}.
\newblock \emph{The International Journal of Robotics Research}, 36\penalty0 (9):\penalty0 1000--1021, 2017.

\bibitem[Andersson et~al.(2019)Andersson, Gillis, Horn, Rawlings, and Diehl]{andersson2019casadi}
Joel~AE Andersson, Joris Gillis, Greg Horn, James~B Rawlings, and Moritz Diehl.
\newblock \href{https://link.springer.com/article/10.1007/s12532-018-0139-4}{CasADi: a software framework for nonlinear optimization and optimal control}.
\newblock \emph{Mathematical Programming Computation}, 11:\penalty0 1--36, 2019.

\bibitem[Dubanchet et~al.(2015)Dubanchet, Saussi{\'e}, Alazard, B{\'e}rard, and Peuv{\'e}dic]{dubanchet2015modeling}
Vincent Dubanchet, David Saussi{\'e}, Daniel Alazard, Caroline B{\'e}rard, and Catherine~Le Peuv{\'e}dic.
\newblock \href{https://link.springer.com/article/10.1007/s12567-015-0082-4}{Modeling and control of a space robot for active debris removal}.
\newblock \emph{CEAS Space Journal}, 7\penalty0 (2):\penalty0 203--218, 2015.

\bibitem[Farivarnejad et~al.(2021)Farivarnejad, Lafmejani, and Berman]{farivarnejad2021fully}
Hamed Farivarnejad, Amir~Salimi Lafmejani, and Spring Berman.
\newblock \href{https://ieeexplore.ieee.org/abstract/document/9438133}{Fully decentralized controller for multi-robot collective transport in space applications}.
\newblock In \emph{2021 IEEE Aerospace Conference (50100)}, pages 1--9. IEEE, 2021.

\bibitem[Flores-Abad et~al.(2014)Flores-Abad, Ma, Pham, and Ulrich]{flores2014review}
Angel Flores-Abad, Ou~Ma, Khanh Pham, and Steve Ulrich.
\newblock \href{https://www.sciencedirect.com/science/article/abs/pii/S0376042114000347}{A review of space robotics technologies for on-orbit servicing}.
\newblock \emph{Progress in aerospace sciences}, 68:\penalty0 1--26, 2014.

\bibitem[Gilpin et~al.(2020)Gilpin, Kurtz, and Lin]{gilpin2020smooth}
Yann Gilpin, Vince Kurtz, and Hai Lin.
\newblock \href{https://ieeexplore.ieee.org/abstract/document/9114883}{A smooth robustness measure of signal temporal logic for symbolic control}.
\newblock \emph{IEEE Control Systems Letters}, 5\penalty0 (1):\penalty0 241--246, 2020.

\bibitem[Girard(2005)]{girard2005reachability}
Antoine Girard.
\newblock Reachability of uncertain linear systems using zonotopes.
\newblock In \emph{International workshop on hybrid systems: Computation and control}, pages 291--305. Springer, 2005.

\bibitem[{Gurobi Optimization, LLC}(2024)]{gurobi}
{Gurobi Optimization, LLC}.
\newblock {Gurobi Optimizer Reference Manual}, 2024.
\newblock URL \url{https://www.gurobi.com}.

\bibitem[Han et~al.(2021)Han, Luo, and Zong]{han2021cooperative}
Nan Han, Jianjun Luo, and Lijun Zong.
\newblock \href{https://ieeexplore.ieee.org/abstract/document/9536397}{Cooperative game method for on-orbit substructure transportation using modular robots}.
\newblock \emph{IEEE Transactions on Aerospace and Electronic Systems}, 58\penalty0 (2):\penalty0 1161--1175, 2021.

\bibitem[Hu et~al.(2024)Hu, Zhang, Wen, and Hu]{hu2024orbit}
Yunhao Hu, Wei Zhang, Hao Wen, and Haiyan Hu.
\newblock \href{https://www.sciencedirect.com/science/article/pii/S0273117724007294?casa_token=ZJkqg8C-iwAAAAAA:R6SIYeQa8UWX4fCLmPeEJpr7M7RB__rEg13fzZgHpdcpITwDk2NZxZzaCWd63lhXt8_Ivw2PXR4}{On-orbit transportation with non-conserved momenta by cooperative space robots}.
\newblock \emph{Advances in Space Research}, 74\penalty0 (10):\penalty0 5179--5191, 2024.

\bibitem[Huang et~al.(2016)Huang, Wang, Meng, Zhang, and Liu]{huang2016impact}
Panfeng Huang, Dongke Wang, Zhongjie Meng, Fan Zhang, and Zhengxiong Liu.
\newblock \href{https://ieeexplore.ieee.org/abstract/document/7470416?casa_token=4Ft5rBPLpBAAAAAA:2kpaLzPUzxFqS-e5GBcqLF9VXAnN6zkMlhWtSeqvHoSYfNqPTI93hCsFFr8j16GtxRlBU7dPYJY}{Impact dynamic modeling and adaptive target capturing control for tethered space robots with uncertainties}.
\newblock \emph{IEEE/ASME Transactions on Mechatronics}, 21\penalty0 (5):\penalty0 2260--2271, 2016.

\bibitem[Jaekel et~al.(2018)Jaekel, Lampariello, Rackl, De~Stefano, Oumer, Giordano, Porges, Pietras, Brunner, Ratti, et~al.]{jaekel2018design}
Steffen Jaekel, Roberto Lampariello, Wolfgang Rackl, Marco De~Stefano, Nassir Oumer, Alessandro~M Giordano, Oliver Porges, Markus Pietras, Bernhard Brunner, John Ratti, et~al.
\newblock \href{https://www.frontiersin.org/articles/10.3389/frobt.2018.00100}{Design and operational elements of the robotic subsystem for the e. deorbit debris removal mission}.
\newblock \emph{Frontiers in Robotics and AI}, 5:\penalty0 100, 2018.

\bibitem[Jankowski et~al.(2024)Jankowski, Mari{\'c}, and Calinon]{jankowski2024airlihockey}
Julius Jankowski, Ante Mari{\'c}, and Sylvain Calinon.
\newblock \href{https://arxiv.org/pdf/2401.14964}{AiRLIHockey: Highly Reactive Contact Control and Stochastic Optimal Shooting}.
\newblock \emph{arXiv preprint arXiv:2401.14964}, 2024.

\bibitem[Landry et~al.(2013)Landry, Dussault, and Mahey]{landry2013heuristic}
Jean-Fran{\c{c}}ois Landry, Jean-Pierre Dussault, and Philippe Mahey.
\newblock \href{https://ieeexplore.ieee.org/abstract/document/6651845}{A heuristic-based planner and improved controller for a two-layered approach for the game of billiards}.
\newblock \emph{IEEE Transactions on Computational Intelligence and AI in Games}, 5\penalty0 (4):\penalty0 325--336, 2013.

\bibitem[Li et~al.(2022)Li, Zhong, Zhu, Xu, Tang, and Zhan]{li2022survey}
Delun Li, Lou Zhong, Wei Zhu, Zhipeng Xu, Qirong Tang, and Wenhao Zhan.
\newblock \href{https://spj.science.org/doi/full/10.34133/2022/9849170}{A survey of space robotic technologies for on-orbit assembly}.
\newblock \emph{Space: Science \& Technology}, 2022.

\bibitem[Maler and Nickovic(2004)]{maler2004monitoring}
Oded Maler and Dejan Nickovic.
\newblock \href{https://link.springer.com/chapter/10.1007/978-3-540-30206-3_12}{Monitoring temporal properties of continuous signals}.
\newblock In \emph{International symposium on formal techniques in real-time and fault-tolerant systems}, pages 152--166. Springer, 2004.

\bibitem[Marcucci et~al.(2023)Marcucci, Petersen, von Wrangel, and Tedrake]{marcucci2023motion}
Tobia Marcucci, Mark Petersen, David von Wrangel, and Russ Tedrake.
\newblock \href{https://www.science.org/doi/10.1126/scirobotics.adf7843}{Motion planning around obstacles with convex optimization}.
\newblock \emph{Science robotics}, 8\penalty0 (84):\penalty0 eadf7843, 2023.

\bibitem[Nenchev and Yoshida(1999)]{nenchev1999impact}
Dragomir~N Nenchev and Kazuya Yoshida.
\newblock \href{https://ieeexplore.ieee.org/abstract/document/768186}{Impact analysis and post-impact motion control issues of a free-floating space robot subject to a force impulse}.
\newblock \emph{IEEE Transactions on Robotics and Automation}, 15\penalty0 (3):\penalty0 548--557, 1999.

\bibitem[Nierhoff et~al.(2015)Nierhoff, Leibrandt, Lorenz, and Hirche]{nierhoff2015robotic}
Thomas Nierhoff, Konrad Leibrandt, Tamara Lorenz, and Sandra Hirche.
\newblock \href{https://ieeexplore.ieee.org/abstract/document/7210189}{Robotic billiards: understanding humans in order to counter them}.
\newblock \emph{IEEE transactions on cybernetics}, 46\penalty0 (8):\penalty0 1889--1899, 2015.

\bibitem[Pant et~al.(2018)Pant, Abbas, Quaye, and Mangharam]{pant2018fly}
Yash~Vardhan Pant, Houssam Abbas, Rhudii~A Quaye, and Rahul Mangharam.
\newblock \href{https://core.ac.uk/download/pdf/214208523.pdf}{Fly-by-logic: Control of multi-drone fleets with temporal logic objectives}.
\newblock In \emph{2018 ACM/IEEE 9th International Conference on Cyber-Physical Systems (ICCPS)}, pages 186--197. IEEE, 2018.

\bibitem[Paraskevas et~al.(2015)Paraskevas, Mitros, and Papadopoulos]{paraskevas2015inertia}
I~Paraskevas, Zisos Mitros, and Evangelos Papadopoulos.
\newblock \href{https://nereus.mech.ntua.gr/Documents/pdf_ps/astra15-2.pdf}{On Inertia and Stiffness Effects During Impact Docking}.
\newblock In \emph{13th Symp. on Advanced Space Technology in Robotics and Automation,(ASTRA ‘15)}, 2015.

\bibitem[Qian et~al.(2016)Qian, Navarro, de~La~Fortelle, and Moutarde]{qian2016motion}
Xiangjun Qian, I{\~n}aki Navarro, Arnaud de~La~Fortelle, and Fabien Moutarde.
\newblock \href{https://hal.science/hal-01425647/}{Motion planning for urban autonomous driving using B{\'e}zier curves and MPC}.
\newblock In \emph{2016 IEEE 19th international conference on intelligent transportation systems (ITSC)}, pages 826--833. Ieee, 2016.

\bibitem[Raman et~al.(2014)Raman, Maasoumy, and Donz{\'e}]{raman2014model}
Vasumathi Raman, Mehdi Maasoumy, and Alexandre Donz{\'e}.
\newblock \href{https://dl.acm.org/doi/abs/10.1145/2593458.2593472}{Model predictive control from signal temporal logic specifications: A case study}.
\newblock In \emph{Proceedings of the 4th ACM SIGBED International Workshop on Design, Modeling, and Evaluation of Cyber-Physical Systems}, pages 52--55, 2014.

\bibitem[Roque et~al.(2025)Roque, Phodapol, Krantz, Lim, Verhagen, Jiang, D{\"o}rner, Mao, Tibert, Siegwart, et~al.]{roque2025towards}
Pedro Roque, Sujet Phodapol, Elias Krantz, Jaeyoung Lim, Joris Verhagen, Frank~J Jiang, David D{\"o}rner, Huina Mao, Gunnar Tibert, Roland Siegwart, et~al.
\newblock \href{https://arxiv.org/abs/2501.16973}{Towards Open-Source and Modular Space Systems with ATMOS}.
\newblock \emph{arXiv preprint arXiv:2501.16973}, 2025.

\bibitem[Sadraddini and Belta(2015)]{sadraddini2015robust}
Sadra Sadraddini and Calin Belta.
\newblock \href{https://ieeexplore.ieee.org/abstract/document/7447084}{Robust temporal logic model predictive control}.
\newblock In \emph{2015 53rd Annual Allerton Conference on Communication, Control, and Computing (Allerton)}, pages 772--779. IEEE, 2015.

\bibitem[Smith et~al.(2016)Smith, Barlow, Bualat, Fong, Provencher, Sanchez, and Smith]{smith2016astrobee}
Trey Smith, Jonathan Barlow, Maria Bualat, Terrence Fong, Christopher Provencher, Hugo Sanchez, and Ernest Smith.
\newblock \href{https://ntrs.nasa.gov/api/citations/20160007769/downloads/20160007769.pdf}{Astrobee: A new platform for free-flying robotics on the international space station}.
\newblock In \emph{International Symposium on Artificial Intelligence, Robotics, and Automation in Space (i-SAIRAS)}, number ARC-E-DAA-TN31584, 2016.

\bibitem[Sun et~al.(2022)Sun, Chen, Mitra, and Fan]{sun2022multi}
Dawei Sun, Jingkai Chen, Sayan Mitra, and Chuchu Fan.
\newblock \href{https://ieeexplore.ieee.org/document/9696363}{Multi-agent motion planning from signal temporal logic specifications}.
\newblock \emph{IEEE Robotics and Automation Letters}, 7\penalty0 (2):\penalty0 3451--3458, 2022.

\bibitem[Tagliabue et~al.(2019)Tagliabue, Kamel, Siegwart, and Nieto]{tagliabue2019robust}
Andrea Tagliabue, Mina Kamel, Roland Siegwart, and Juan Nieto.
\newblock \href{https://journals.sagepub.com/doi/full/10.1177/0278364919854131}{Robust collaborative object transportation using multiple MAVs}.
\newblock \emph{The International Journal of Robotics Research}, 38\penalty0 (9):\penalty0 1020--1044, 2019.

\bibitem[Uyama et~al.(2012)Uyama, Fujii, Nagaoka, and Yoshida]{uyama2012experimental}
N~Uyama, Y~Fujii, K~Nagaoka, and K~Yoshida.
\newblock \href{http://www.mech.kyutech.ac.jp/srl/papers/2012isairas_uyama.pdf}{Experimental evaluation of contact-impact dynamics between a space robot with a compliant wrist and a free-flying object}.
\newblock In \emph{International symposium on artificial intelligence, robotics and automation in space}, 2012.

\bibitem[Verhagen et~al.(2024)Verhagen, Lindemann, and Tumova]{verhagen2024temporally}
Joris Verhagen, Lars Lindemann, and Jana Tumova.
\newblock \href{https://ieeexplore.ieee.org/abstract/document/10644216}{Temporally robust multi-agent stl motion planning in continuous time}.
\newblock In \emph{2024 American Control Conference (ACC)}, pages 251--258. IEEE, 2024.

\bibitem[von Wrangel and Tedrake(2024)]{von2024using}
David von Wrangel and Russ Tedrake.
\newblock \href{https://ieeexplore.ieee.org/abstract/document/10802426}{Using Graphs of Convex Sets to Guide Nonconvex Trajectory Optimization}.
\newblock In \emph{2024 IEEE/RSJ International Conference on Intelligent Robots and Systems (IROS)}, pages 9863--9870. IEEE, 2024.

\bibitem[Zermane et~al.(2024)Zermane, Moussafir, Yan, and Kheddar]{zermane2024minimal}
Ahmed Zermane, L{\'e}o Moussafir, Youcan Yan, and Abderrahmane Kheddar.
\newblock \href{https://ieeexplore.ieee.org/abstract/document/10742560}{Minimal Impact Pokes to Place Objects on Planar Surfaces}.
\newblock \emph{IEEE Robotics and Automation Letters}, 2024.

\bibitem[Zhang et~al.(2015)Zhang, Jia, Chen, and Sun]{zhang2015pre}
Long Zhang, Qingxuan Jia, Gang Chen, and Hanxu Sun.
\newblock \href{https://www.sciencedirect.com/science/article/pii/S1000936115001053}{Pre-impact trajectory planning for minimizing base attitude disturbance in space manipulator systems for a capture task}.
\newblock \emph{Chinese Journal of Aeronautics}, 28\penalty0 (4):\penalty0 1199--1208, 2015.

\end{thebibliography}
